\documentclass[12pt]{article}

\usepackage{etex}

\usepackage{fullpage}

\usepackage{tikz}
\usetikzlibrary{decorations,shapes} 

\usepackage{tikz-cd}
\tikzcdset{
  shift left/.default=+0.7ex,
  shift right/.default=+0.7ex}

\newenvironment{tikzar}[1][]{{}\kern-4pt\begin{tikzcd}[ampersand replacement=\&,#1]}%
{\end{tikzcd}\kern-4pt{}}

\usepackage{enumitem}
\setlist[itemize]{noitemsep, topsep=0pt}
\setlist[enumerate]{noitemsep, topsep=0pt}

\usepackage{amsmath,stmaryrd}
\usepackage{amsfonts,amssymb}
\usepackage{pstricks}
\usepackage{amsthm}
\usepackage{rotating}

\usepackage[LGR,T1]{fontenc}

\newdimen\proofrulebreadth \proofrulebreadth=.05em
\newdimen\proofdotseparation \proofdotseparation=1.25ex
\newdimen\proofrulebaseline \proofrulebaseline=2ex
\newcount\proofdotnumber \proofdotnumber=3
\let\then\relax
\def\hfi{\hskip0pt plus.0001fil}
\mathchardef\squigto="3A3B
\newif\ifinsideprooftree\insideprooftreefalse
\newif\ifonleftofproofrule\onleftofproofrulefalse
\newif\ifproofdots\proofdotsfalse
\newif\ifdoubleproof\doubleprooffalse
\let\wereinproofbit\relax
\newdimen\shortenproofleft
\newdimen\shortenproofright
\newdimen\proofbelowshift
\newbox\proofabove
\newbox\proofbelow
\newbox\proofrulename
\def\shiftproofbelow{\let\next\relax\afterassignment\setshiftproofbelow\dimen0 }
\def\shiftproofbelowneg{\def\next{\multiply\dimen0 by-1 }%
\afterassignment\setshiftproofbelow\dimen0 }
\def\setshiftproofbelow{\next\proofbelowshift=\dimen0 }
\def\setproofrulebreadth{\proofrulebreadth}

\def\prooftree{
\ifnum  \lastpenalty=1
\then   \unpenalty
\else   \onleftofproofrulefalse
\fi
\ifonleftofproofrule
\else   \ifinsideprooftree
        \then   \hskip.5em plus1fil
        \fi
\fi
\bgroup
\setbox\proofbelow=\hbox{}\setbox\proofrulename=\hbox{}%
\let\justifies\proofover\let\leadsto\proofoverdots\let\Justifies\proofoverdbl
\let\using\proofusing\let\[\prooftree
\ifinsideprooftree\let\]\endprooftree\fi
\proofdotsfalse\doubleprooffalse
\let\thickness\setproofrulebreadth
\let\shiftright\shiftproofbelow \let\shift\shiftproofbelow
\let\shiftleft\shiftproofbelowneg
\let\ifwasinsideprooftree\ifinsideprooftree
\insideprooftreetrue
\setbox\proofabove=\hbox\bgroup$\displaystyle 
\let\wereinproofbit\prooftree
\shortenproofleft=0pt \shortenproofright=0pt \proofbelowshift=0pt
\onleftofproofruletrue\penalty1
}

\def\eproofbit{
\ifx    \wereinproofbit\prooftree
\then   \ifcase \lastpenalty
        \then   \shortenproofright=0pt  
        \or     \unpenalty\hfil         
        \or     \unpenalty\unskip       
        \else   \shortenproofright=0pt  
        \fi
\fi
\global\dimen0=\shortenproofleft
\global\dimen1=\shortenproofright
\global\dimen2=\proofrulebreadth
\global\dimen3=\proofbelowshift
\global\dimen4=\proofdotseparation
\global\count255=\proofdotnumber
$\egroup  
\shortenproofleft=\dimen0
\shortenproofright=\dimen1
\proofrulebreadth=\dimen2
\proofbelowshift=\dimen3
\proofdotseparation=\dimen4
\proofdotnumber=\count255
}

\def\proofover{
\eproofbit 
\setbox\proofbelow=\hbox\bgroup 
\let\wereinproofbit\proofover
$\displaystyle
}%
\def\proofoverdbl{
\eproofbit 
\doubleprooftrue
\setbox\proofbelow=\hbox\bgroup 
\let\wereinproofbit\proofoverdbl
$\displaystyle
}%
\def\proofoverdots{
\eproofbit 
\proofdotstrue
\setbox\proofbelow=\hbox\bgroup 
\let\wereinproofbit\proofoverdots
$\displaystyle
}%
\def\proofusing{
\eproofbit 
\setbox\proofrulename=\hbox\bgroup 
\let\wereinproofbit\proofusing
\kern0.3em$
}

\def\endprooftree{
\eproofbit 
  \dimen5 =0pt
\dimen0=\wd\proofabove \advance\dimen0-\shortenproofleft
\advance\dimen0-\shortenproofright
\dimen1=.5\dimen0 \advance\dimen1-.5\wd\proofbelow
\dimen4=\dimen1
\advance\dimen1\proofbelowshift \advance\dimen4-\proofbelowshift
\ifdim  \dimen1<0pt
\then   \advance\shortenproofleft\dimen1
        \advance\dimen0-\dimen1
        \dimen1=0pt
        \ifdim  \shortenproofleft<0pt
        \then   \setbox\proofabove=\hbox{%
                        \kern-\shortenproofleft\unhbox\proofabove}%
                \shortenproofleft=0pt
        \fi
\fi
\ifdim  \dimen4<0pt
\then   \advance\shortenproofright\dimen4
        \advance\dimen0-\dimen4
        \dimen4=0pt
\fi
\ifdim  \shortenproofright<\wd\proofrulename
\then   \shortenproofright=\wd\proofrulename
\fi
\dimen2=\shortenproofleft \advance\dimen2 by\dimen1
\dimen3=\shortenproofright\advance\dimen3 by\dimen4
\ifproofdots
\then
        \dimen6=\shortenproofleft \advance\dimen6 .5\dimen0
        \setbox1=\vbox to\proofdotseparation{\vss\hbox{$\cdot$}\vss}%
        \setbox0=\hbox{%
                \advance\dimen6-.5\wd1
                \kern\dimen6
                $\vcenter to\proofdotnumber\proofdotseparation
                        {\leaders\box1\vfill}$%
                \unhbox\proofrulename}%
\else   \dimen6=\fontdimen22\the\textfont2 
        \dimen7=\dimen6
        \advance\dimen6by.5\proofrulebreadth
        \advance\dimen7by-.5\proofrulebreadth
        \setbox0=\hbox{%
                \kern\shortenproofleft
                \ifdoubleproof
                \then   \hbox to\dimen0{%
                        $\mathsurround0pt\mathord=\mkern-6mu%
                        \cleaders\hbox{$\mkern-2mu=\mkern-2mu$}\hfill
                        \mkern-6mu\mathord=$}%
                \else   \vrule height\dimen6 depth-\dimen7 width\dimen0
                \fi
                \unhbox\proofrulename}%
        \ht0=\dimen6 \dp0=-\dimen7
\fi
\let\doll\relax
\ifwasinsideprooftree
\then   \let\VBOX\vbox
\else   \ifmmode\else$\let\doll=$\fi
        \let\VBOX\vcenter
\fi
\VBOX   {\baselineskip\proofrulebaseline \lineskip.2ex
        \expandafter\lineskiplimit\ifproofdots0ex\else-0.6ex\fi
        \hbox   spread\dimen5   {\hfi\unhbox\proofabove\hfi}%
        \hbox{\box0}%
        \hbox   {\kern\dimen2 \box\proofbelow}}\doll%
\global\dimen2=\dimen2
\global\dimen3=\dimen3
\egroup 
\ifonleftofproofrule
\then   \shortenproofleft=\dimen2
\fi
\shortenproofright=\dimen3
\onleftofproofrulefalse
\ifinsideprooftree
\then   \hskip.5em plus 1fil \penalty2
\fi
}

\usepackage{hyperref}

\newcommand{\id}{{\rm id}}

\newcommand{\CCC}{{\cal C}}

\mathcode`\<="4268 
\mathcode`\>="5269 
\mathchardef\gt="313E 
\mathchardef\lt="313C 
\newsavebox{\barr}
\savebox{\barr}{\hspace*{-9.5pt}\raisebox{1.25pt}{$
\scriptscriptstyle%
|$}\hspace*{4.5pt}} 
\newsavebox{\barrleft}
\savebox{\barrleft}{\hspace*{-8.5pt}\raisebox{1.25pt}{$
\scriptscriptstyle%
|$}\hspace*{10pt}}

 \def\pushright#1{{
    \parfillskip=0pt            
    \widowpenalty=10000         
    \displaywidowpenalty=10000  
    \finalhyphendemerits=0      
    \leavevmode                 
    \unskip                     
    \nobreak                    
    \hfil                       
    \penalty50                  
    \hskip.2em                  
    \null                       
    \hfill                      
    {#1}                        
    \par}}                      

 \def\qed{\pushright{$\square$}\penalty-700 \smallskip}

\newenvironment{prf}[1]{\begin{trivlist} \item[{\bf ~Proof}#1.]}%
{\qed\end{trivlist}}

\newcommand{\beq}{\begin{equation}}
\newcommand{\eeq}{\end{equation}}
\newcommand{\ba}[1]{\begin{array}{#1}}
\newcommand{\ea}{\end{array}}
\newcommand{\bea}{\begin{eqnarray}}
\newcommand{\eea}{\end{eqnarray}}
\newcommand{\bear}{\begin{eqnarray*}}
\newcommand{\eear}{\end{eqnarray*}}
\newcommand{\bpr}{\begin{prf}{}}
\newcommand{\epr}{\end{prf}}
\newcommand{\bprf}[1]{\begin{prf}{#1}}
\newcommand{\eprf}{\end{prf}}

\theoremstyle{plain}

\theoremstyle{remark}

\renewcommand{\to}{\longrightarrow}

\newcommand{\tto}[1]{\xrightarrow{#1}}

\newcommand{\oot}[1]{\xleftarrow{#1}}

\pgfdeclaredecoration{single line}{initial}{
        \state{initial}[width=\pgfdecoratedpathlength-1sp]{\pgfpathmoveto{\pgfpointorigin}}
        \state{final}{\pgfpathlineto{\pgfpointorigin}}
}

\tikzset{
        raise line/.style={
                decoration={single line, raise=#1}, decorate
        }
}

\renewcommand{\paragraph}[1]{\vspace{.3\baselineskip}\noindent\textbf{#1}}

\newcommand{\state}[2]{\left[{#1},{#2}\right]}

\newcommand{\undefined}[1]{{#1}\!\!\uparrow}

\newcommand{\DP}{\Omega}
\newcommand{\xplan}{P}
\newcommand{\explains}{prediction}
\newcommand{\Params}{Y}
\newcommand{\Paramss}{X}
\newcommand{\Causess}{{\rm Causes}}
\newcommand{\Effectss}{{\rm Effects}}
\newcommand{\causation}{{causal process}}
\newcommand{\causations}{{causal processes}}

\newcommand{\ttimes}{\otimes}

\newcommand{\cmn}{\Delta}
\newcommand{\cun}{\top}

\newcommand{\comp}[2]{{#2}\circ {#1}}

\newcommand{\uev}[1]{\Semantics{#1}}

\newcommand{\Semantics}[1]{\llbracket{#1}\rrbracket}

\graphicspath{{PICS/}}

\begin{document}

\title{Causality and deceit:\\ 
Do androids watch action movies?}
\author{Dusko~Pavlovic\thanks{Partially supported by NSF and AFOSR.} \and Temra Pavlovic
	}

\date{}

\maketitle
\begin{abstract} 
\noindent We seek causes through science,  religion, and in everyday life. We get excited when a big rock causes a big splash, and we get scared when it tumbles without a cause. But our causal cognition is usually biased.  The \emph{why}\/ is influenced by the \emph{who}. It is influenced by the \emph{self}, and by \emph{others}. We share rituals, we watch  action movies, and we influence each other to believe in the same causes. Human mind is packed with subjectivity because shared cognitive biases bring us together. But they also make us vulnerable.

An artificial mind is deemed to be more objective than the human mind. After many years of science-fiction fantasies about even-minded androids, they are now sold as personal or expert assistants, as brand advocates, as policy or candidate supporters, as network influencers.  Artificial agents have been stunningly successful in disseminating artificial causal beliefs among humans. As malicious  artificial agents continue to manipulate human cognitive biases, and deceive human communities into ostensive but expansive causal illusions, the hope for defending us has been vested into developing benevolent artificial agents, tasked with preventing and mitigating cognitive distortions inflicted upon us by their malicious cousins. Can the distortions of human causal cognition be corrected on a more solid foundation of artificial causal cognition?

In the present paper, we study a simple model of causal cognition, viewed as a quest for causal models. We show that, under very mild and hard to avoid assumptions, there are always self-confirming causal models, which perpetrate self-deception, and seem to preclude a royal road to objectivity. 
\end{abstract}

\newpage
\tableofcontents	

\clearpage

\section{Introduction: Causal cognition and its vulnerabilities}
\label{Sec:Intro}

\subsection{Causal cognition in life and science}\label{Sec:IntroOne}
Causal cognition drives the interactions of organisms and organizations with their environments at many levels.  At the level of perception, spatial and temporal correlations are often interpreted as causations, and directly used for predictions \cite{Michotte}. At the level of science, testing causal hypotheses is an important part of the practice of experiment design \cite{fisher1935design}, although the general concept of causation seldom explicitly occurs in scientific theories \cite{RussellB:cause}. 
It does occur, of course, quite explicitly in metaphysics and in theory of science \cite{bunge2012causality}, and most recently in AI\footnote{\emph{AI}\/ is usually interpreted as the acronym of \emph{Artificial Intelligence}. But the discipline given that name in 1956 by John McCarthy evolved in the meantime in many directions, some of which are hard to relate with any of the usual meanings of the word "intelligence". We are thus forced to either keep updating the concept of "intelligence" to match the artificial versions, or to unlink the term "AI" from its etymology, and to use it as a word rather than as an acronym, like it happened with the words gif, captcha, gulag, or snafu. The latter choice seems preferable, at least in the present context.}\cite{pearl2009causality,spirtes2000causation}. 


\subsection{Causal cognition and launching effects}
Causal theories may or may not permeate science (depending on how you look at it) but they certainly permeate life. Why did the ball bounce? Why did my program crash? Why did the boat explode? Who caused the accident? Why did the chicken cross the street? We need to know. We seek to debug programs, to prevent accidents, and to understand chicken behaviors. We often refuse to move on without an explanation. Lions move on, mice run away, but humans want to understand, to predict, to control.

In some cases, the path to a causal hypothesis and the method to test it are straightforward. When a program crashes, the programmer follows its execution path. The cause of the crash must be on it. The causal hypothesis is tested by eliminating the cause and verifying that the effect has been eliminated as well. 
\begin{figure}[ht!]
\begin{center}
\includegraphics[height=3.8cm]{newton.epsf}
\hspace{3em}\includegraphics[height=3.8cm]{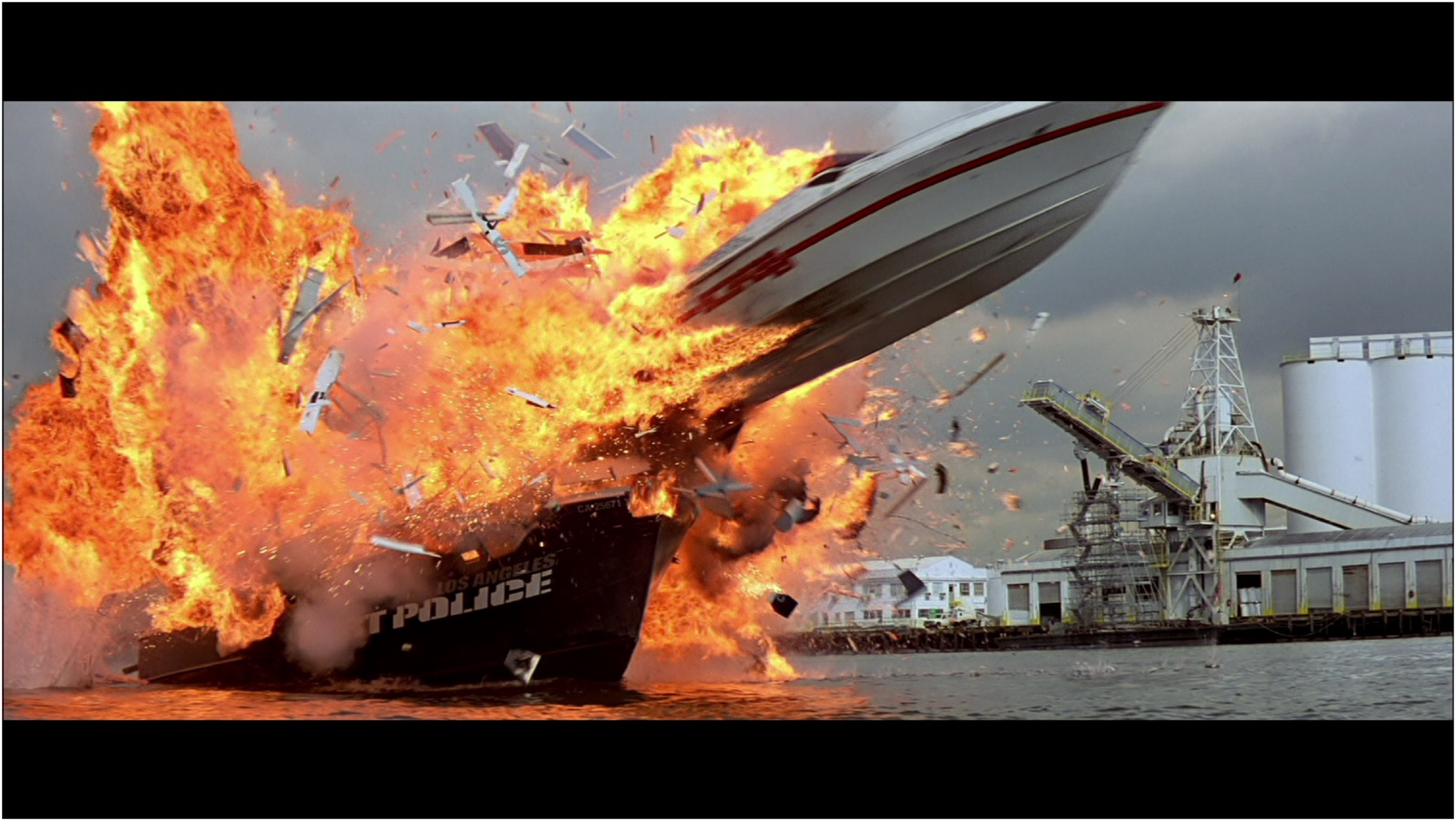}
\caption{Launching effects}
\label{Fig:boats}
\end{center}
\end{figure}
If a ball bounces like in Newton's cradle, in Fig.~\ref{Fig:boats} on the left, we see that the ball was launched by another ball. If a boat explodes like in Fig.~\ref{Fig:boats} on the right, we see that the explosion was caused by another boat crashing. Action movies are chains of causes and effects, packed between a problem and a solution, which are usually presented as a big cause and a big effect.

In other cases, establishing a causal hypothesis may be hard. It is obvious that the collision caused the explosion; but who caused the collision? It is obvious that the bouncing ball extends the trajectory of the hitting ball; but how is the force transferred from the hitting ball to the bouncing ball through all the balls in-between, that don't budge at all?

Such questions drive our quest for causes, and our fascination with illusions. They drive us into science, and into magic. In a physics lab, a physics teacher would explain the physical law behind Newton's cradle. But in a magic lab, a magic teacher would present a magic cradle, looking exactly the same like Newton's cradle, but behaving differently. One time the hitting ball hits, and the bouncing ball bounces. Another time the hitting ball hits, but the bouncing ball does not bounce. But later, unexpectedly, the bouncing ball bounces all on its own, without the hitting ball hitting. 
Still later, the bouncing ball bounces again, and the hitting ball hits \emph{after}\/ that. Magic! Gradually you understand that the magic cradle is controlled by the magic teacher: he makes the balls bounce or stick at his will. 

\begin{figure}[h!t]
\begin{center}
\includegraphics[height=4cm]{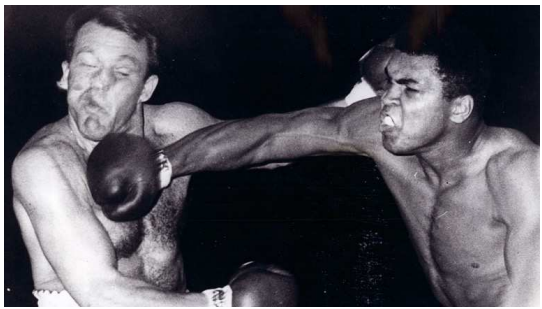}\qquad \qquad
\includegraphics[height=4cm]{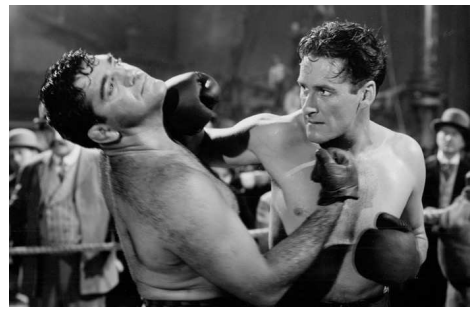}
\caption{The cause of an effect may appear obvious, but the appearance may deceive}
\label{Fig:ali}
\end{center}
\end{figure}

Michotte studied such \emph{"launching effects"}\/  in his seminal book \cite{Michotte}. The action movie industry is built upon a wide range of techniques for producing such effects. While the example in Fig.~\ref{Fig:boats} on the right requires pyrotechnic engineering, the example in Fig.~\ref{Fig:ali} on the right is close to Michotte's lab setups. 
Observing subtle details, we can usually tell apart a real effect from an illusion. But illusions are often more exciting, or less costly, and we accept them. We enjoy movies, magic, superstition. We abhor randomness and complexity, and seek the simplest causal explanations. We like to be deceived, and we deceive ourselves.

%
%

\subsection{Causal cognition as a security problem}\label{Sec:security}
Cognitive bias and belief perseverance are familiar and well-studied properties of human cognition, promoted in evolution because they strengthen the group cohesion  \cite{CogBias-handbook}. As human interactions increasingly spread through media, human beliefs are hijacked, their biases are manipulated, and social cohesion is amplified and abused at ever larger scales. 
This is also well-known and well-studied in psychology \cite[Part~VI]{buss2015handbook-2} and in social sciences \cite{zuboff2019age}. However, on the information technology side, cognitive bias manipulations on the web market and in cyberspace have been so profitable in practice, that there was no time for theory.

%
%
%
%
We live in times of advertising and publicity campaigns. Brands and crowds are built and steered using the same tools, and from the same materials: human cognition and sensitivities. The AI tools, recently added to the social engineering toolkit, appear to be among the most effective ones, and among the least understood. The practice is far ahead of theory. The research reported in this paper is a part of a broader effort \cite{PavlovicD:CathyFest,PavlovicD:NSPW11,PavlovicD:HoTSoS15,PavlovicD:ICDCIT12,PavlovicD:AMAI17} towards developing models, methods, and tools for this area, where human and computational behaviors are not just inseparable, but they determine each other; and they do not just determine each other, but they seek  to control each other.

The specific motivating question that we pursue is: \emph{Is artificial causal cognition suitable as a tool to defend human causal cognition from the rapid strategic advances in cognitive bias manipulation, influencing, and deceit?}

There are, of course, many aspects of this question, and many approaches to it. The combined backgrounds of this paper's authors provide equipment only for scaling the south-west ridge: where the west cliff of \emph{computation}\/ meets the south cliff of \emph{communication}. Neither of us is equipped to place or interpret our model and results in the context of the extant research in psychology, or social sciences, where they should be empirically validated. 
We are keen to present it to a mixed audience hoping that there are gates between the silos. Rather than attempt to shed light on the problem from a direction that is unfamiliar to us, we are hoping to shed light on the problem from a direction that is unfamiliar to the reader, while doing our honest best to avoid throwing any avoidable shadows. If the reader sheds some light from their direction, this multi-dimensional problem may emerge in its multi-disciplinary space.

\section{Background: Causality theories and models}
\label{Sec:Causality}

Causality is studied in different communities from different standpoints. The following high level overview is mainly intended to put us at a starting position. The reader is welcome to fast-forward at any point.

\subsection{A very brief history of causality}\label{Sec:history}
Causal relations impose order on the events in the world: an event $a$ causes an event $b$, which causes an event $c$, whereas the event $d$ is unrelated to $a$ or $b$, but may contribute to $c$. 
An event may have multiple causes, and multiple effects, and they may all be distant from one another in time and in space.  We impose order on the world by thinking in terms of causes and effects.

But connecting causes and effects also causes difficulties. In Fig.~\ref{Fig:boats}, the force on one end of Newton's cradle causes the effect on the other end without affecting anything in-between; whereas the boats collide and are subsequently engulfed in the explosion, but the explosion is not caused by the collision, but staged. Tellingly, such decouplings are called \emph{special effects}. Our eye distrusts the physical effect on the left, and accepts the special effect on the right.  Recognizing such remote causations and adjacent non-causations requires learning about the unobservable causes of observable effects that fill the world, and about the unobservable effects of observable causes, and even about the unobservable causes of unobservable effects. Imposing the causal order on the world forces us to think about the unobservable, whether they are black holes or spirits of our ancestors. 

%
%
%
%
%

While acting upon some causes to prevent some effects is a matter of individual adaptations and survival for humans and androids, plants and animals, understanding the general process of causation has been a matter of civilization and cognition. While the efforts in the latter direction surely go back to the dawn of mankind, the early accounts go back to pre-Socratic philosophy, and have been traditionally subsumed under \emph{Metaphysics}, which was the (post-Socratic) title of Aristotle's manuscript that came \emph{after} (i.e. it was the \emph{"meta-"} to) his Physics \cite{Aristotle:Physics}. We mention just three paradigmatic steps towards the concept of causation:
\begin{enumerate}[label=\roman*), labelindent=\parindent]
\item {Parmenides:}\/ "Nothing comes from nothing." 

\item {Heraclitus:}\/ "Everything has a cause, nothing is its own cause."

\item {Aristotle:}\/ "Everything comes from an {\em Uncaused Cause}."
\end{enumerate}

Step (i) thus introduces the principle of causation; step (ii) precludes causal cycles, and thus introduces the problem of \emph{infinite regression}\/ through causes of causes; and step (iii) resolves this problem by the argument\footnote{"It is clear, then, that though there may be countless instances of the perishing of unmoved movers, and though many things that move themselves perish and are succeeded by others that come into being, and though one thing that is unmoved moves one thing while another moves another, nevertheless there is something that comprehends them all, and that as something apart from each one of them, and this it is that is the cause of the fact that some things are and others are not and of the continuous process of change; and this causes the motion of the other movers, while they are the causes of the motion of other things. Motion, then, being eternal, the First Mover, if there is but one, will be eternal also; if there are more than one, there will be a plurality of such eternal movers." \cite[258b--259a]{Aristotle:Physics}} that in Christian theology came to be interpreted as the \emph{cosmological proof} of existence and uniqueness of  God. Just like the quest for causes of particular phenomena leads to magic and superstition, the quest for a general Uncaused Cause leads to monotheistic religion. This logical pattern persists in modern cosmology, where the Uncaused Cause arises as the initial gravitational singularity, known as the Big Bang. Although it is now described mathematically, it still spawns untestable theories. The singularity can be avoided using mathematical devices, inflating and smoothening the Uncaused Cause into a field; but preventing its untestable implications requires logical devices.

%
%
%
%
%
%

While modern views of a \emph{global}\/ cause of the world may still appear somewhat antique, the modern theories of \emph{local}\/ causations evolved with science. The seed of the modern idea that causes can be separated and made testable through \emph{intervention}\/ \cite{GopnikA:learning,SlomanS:choice,woodward2005making}  was sown by Galileo: "That and no other is to be called cause, at the presence of which the effect always follows, and at whose removal the effect disappears" (cited in \cite{bunge2012causality}). As science progressed, philosophers echoed the same idea in their own words, e.g., "It is necessary to our using the word cause that we should believe not only that the antecedent always has been followed by the consequent, but that as long as the present constitution of things endures, it always will be so" \cite{mill1904system}. 

But David Hume remained unconvinced: "When I see, for instance, a billiard-ball moving in a straight line towards another; even suppose motion in the second ball should by accident be suggested to me, as the result of their contact or impulse; may I not conceive, that a hundred different events might as well follow from the cause? [\ldots] The mind can never possibly find the effect in the supposed cause, by the most accurate scrutiny and examination. For the effect is different from the cause, and can never be discovered in it." \cite[4.10]{Hume:enquiry} Hume's objections famously shook Immanuel Kant from his metaphysical  "dogmatic slumber", and they led him on the path of \emph{"transcendental deduction"}\/ of effects from causes, based on \emph{synthetic a priori}\/ judgements, that are to be found mostly in mathematics \cite{KantI:critique}. Scientists,  however never managed to prove any of their causal hypotheses by pure reason, but took over the world by disproving their hypotheses in lab experiments, whereupon they would always proceed to formulate better causal hypotheses, which they would then try to disprove again, and so on. The imperative of falsifiable hypotheses, and the idea of progress by disproving them empirically, suggested the end of metaphysics, and of causality, and it prompted Bertrand Russell to announce in 1912: "We shall have to give up hope of finding causal laws such as Mill contemplated, as any causal sequence which we have observed may at any moment be falsified without a falsification of any laws of the kind that the actual sciences aim at establishing. [\ldots] All philosophers of every school imagine that causation is one of the fundamental axioms or postulates of science, yet, oddly enough, in advanced science such as [general relativity], the word 'cause' never occurs. [\ldots] The law of causality, I believe, like much that passes muster among philosophers, is a relic of a bygone age, surviving, like the monarchy, only because it is erroneously supposed to do no harm." \cite{RussellB:cause}

\subsection{A very high-level view of causal models}\label{Sec:models}
In spite of the logical difficulties, the concept of causation persisted. General relativity resolved the problem of action at distance, that hampered Newtonian theory of gravitation, by replacing causal interactions by fields; but the causal cones continued to mushroom through the spacetime of special relativity, even along the closed timelike curves. In quantum mechanics, causality was one of the main threads in the discussions between Einstein and Bohr \cite{Bohr51}. It was also the stepping stone into Bohm's quantum dialectical materialism \cite{BohmD:causality,BohmD:letters}. Last but not least, causality remains central in the modern axiomatizations of quantum mechanics \cite{Ciribella,CoeckeB:CQM-caus}, albeit in a weak form. Nevertheless, even the weakest concept of causality implies that the observables must be objective, in the sense that the outcomes of measurements should not depend subjective choices, which precludes covert signaling.

The most influential modern theory of causation arose from Judea Pearl's critique of subjective probability, which led him to \emph{Bayesian networks} \cite{PearlJ:85,pearl2014probabilistic}. As a mathematical structure, a Bayesian net can be thought of as an extension of a Markov chain, where each state is assigned a random variable, and each transition represents a stochastic dependency. Extending Markov chains in this way, of course, induces a completely different interpretation, which is stable only if there are no cycles of state transitions. Hence the causal ordering.  

Originally construed as an AI model, Bayesian networks turned out to be a useful analysis tool in many sciences, from psychology \cite{Gopnik:bayes} to genetics and oncology \cite{KollerD-Segal}. Broad overviews of the theory and of the immense range of its applications can be found in \cite{koller2009probabilistic,pearl2009causality,spirtes2000causation}. Very readable popular accounts are \cite{PearlJ:CACM,pearl2018book}, and \cite{SlomanS:Book} is written from the standpoint of psychology. The causal models imposing the acyclicity requirement on the dependencies between random variables opened opened an alley towards an algorithmic approach to causal reasoning, based on dependency discovery: given some input random variables and some output random variables, specify a network of unobservable random variables whereby the given inputs can cause the given outputs. A good introduction into such discovery algorithms is \cite{spirtes2000causation}. The applications have had a tremendous impact.

A skeptic could, or course, argue that inserting in-between a cause and an effect a causal diagram is just as justified as inserting Kant's synthetic \emph{a priori}\/ judgements. A control theorist could also argue that the canonical language for specifying dependencies between random variables was provided by stochastic calculus a long time ago. They were not presented as elegantly as in bayesian nets, but they were calculated and graphed. On the other hand, the dependencies modeled in control theory are not required to be acyclic. This is, of course, essential, since the acyclicity requirement precludes feedback. Is feedback acausal?  A simple physical system like centrifugal governor\footnote{Centrifugal governor consists of a pair of rotating weights, mounted to open a valve proportionally to their rotation speed, were used to control pressure in steam engines, and before that on the windmill stones. Bayesian nets require an added dimension of time to model such things, basically unfolding the feedback!} obviously obeys the same laws of causation like its feedback-free cousins: an increase in angular momentum causes the valve lever to rise; a wider opening lets more steam out and decreases the pressure; a lower pressure causes a decrease in angular momentum. The only quirk is thus that some physical effects thus cause physical effects on their causes, in continuous time. Whether that is causation or not depends on the modeling framework.

Be it as it may, our very short story about the concepts and the models of causality ends here. The upshot is that \textbf{\emph{there are effective languages for explaining causation}}. The availability and effectiveness of such languages is the central \textbf{assumption} on which the rest of the paper is based. Such languages, effective enough to explain everything, have been provided by metaphysics, and by transcendental deduction, and by bayesian networks, and by stochastic processes.  There are other such languages, and there will be more. The rest of our story proceeds the same for any of them.

\section{Approach: Causes in boxes, tied with strings}\label{Sec:Category}

\subsection{String diagrams}
\begin{figure}[ht]
\medskip
\begin{center}
\begin{minipage}{.3\linewidth}
\begin{center}
\newcommand{\plann}{\includegraphics[height=2.75cm,width=4.05cm]{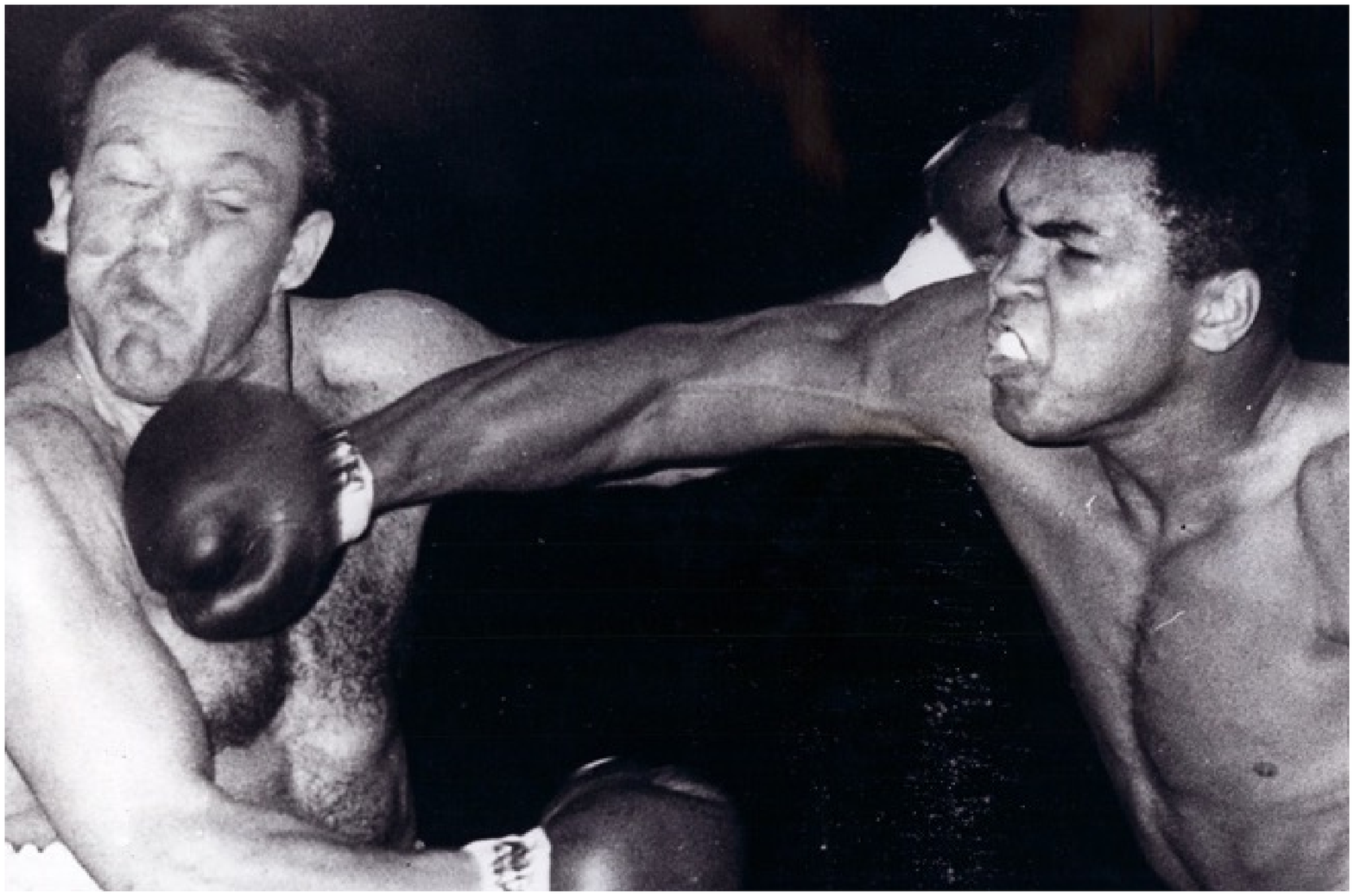}}
\newcommand{\inputtt}{\raisebox{2ex}{\ \bf \large Cause $\rightsquigarrow$  }\includegraphics[height=1cm]{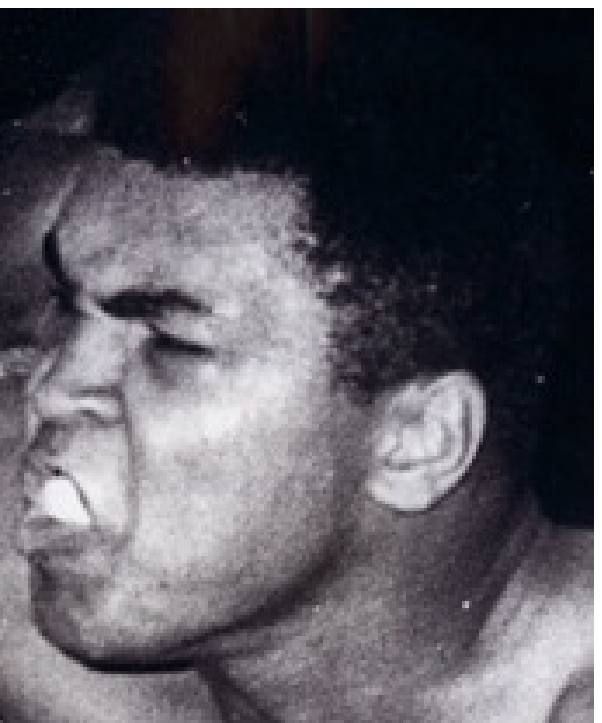}
}
\newcommand{\processt}{\raisebox{2ex}{\large \bf process $\rightsquigarrow$ }}
\newcommand{\outputt}{\raisebox{2ex}{\large \bf Effect $\rightsquigarrow$ }\includegraphics[height=1cm]{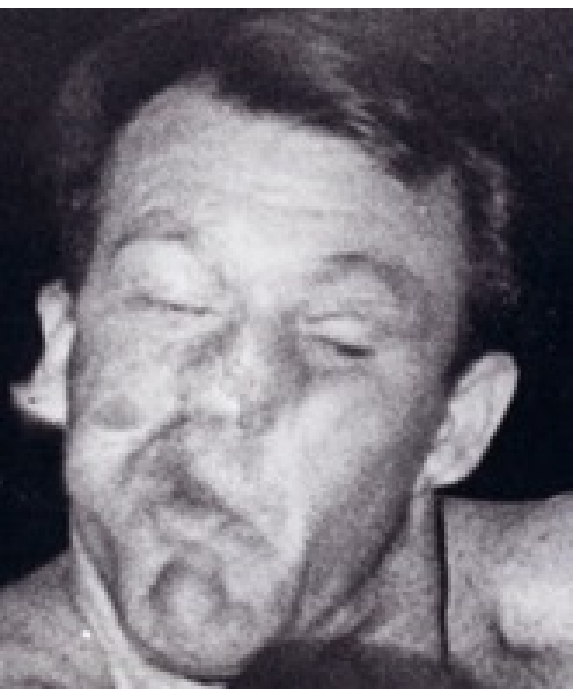}}
\def\JPicScale{.5}
\input{PICS/process.tex}
\end{center}
\end{minipage}
\medskip
\hspace{.25\linewidth}
\begin{minipage}{.3\linewidth}
\newcommand{\plann}{\mbox{\footnotesize \bf course}
}
\newcommand{\xxam}{\begin{minipage}{3.8em}
\footnotesize exam and
\\[-.75ex]
\footnotesize homework\end{minipage}}
\newcommand{\rref}{\begin{minipage}{3em}
\footnotesize reference
\\[-1ex]
\footnotesize request\end{minipage}}
\newcommand{\inputt}{\begin{minipage}{3em}\begin{flushright}
\scriptsize Student
\\
\scriptsize Work\end{flushright}\end{minipage}}
\newcommand{\inputtt}{\begin{minipage}{3em} \begin{flushright}
\scriptsize Course\\
\scriptsize Hardness 
\end{flushright}\end{minipage}}
\newcommand{\inputttt}{\begin{minipage}{3em}\begin{flushright}
\scriptsize Class
\\
\scriptsize Activity\end{flushright}\end{minipage}}
\newcommand{\outputt}{\begin{minipage}{3em}
\scriptsize Grade
\\
\scriptsize Report\end{minipage}}
\newcommand{\recordd}{\mbox{\scriptsize Record}}

\newcommand{\outputtt}{\begin{minipage}{3em}
\scriptsize Reference
\\
\scriptsize Letter\end{minipage}}
\begin{center}
\def\JPicScale{.5}
\ifx\JPicScale\undefined\def\JPicScale{1}\fi
\psset{unit=\JPicScale mm}
\psset{linewidth=0.3,dotsep=1,hatchwidth=0.3,hatchsep=1.5,shadowsize=1,dimen=middle}
\psset{dotsize=0.7 2.5,dotscale=1 1,fillcolor=black}
\psset{arrowsize=1 2,arrowlength=1,arrowinset=0.25,tbarsize=0.7 5,bracketlength=0.15,rbracketlength=0.15}
\begin{pspicture}(0,0)(90,140)
\psline[linewidth=0.75,arrowlength=1.5,arrowinset=0]{<-}(15,47.5)(15,0)
\psline[linewidth=0.75,arrowlength=1.5,arrowinset=0]{<-}(30,140)(30,65)
\pspolygon[linewidth=0.75](0,100)(90,100)(90,40)(0,40)
\rput[r](13.75,10){$\inputtt$}
\rput[l](31.88,130){$\outputt$}
\rput[tl](2.5,96.88){$\plann$}
\psline[linewidth=0.75,arrowlength=1.5,arrowinset=0]{<-}(45,47.5)(45,0)
\psline[linewidth=0.75,arrowlength=1.5,arrowinset=0]{<-}(75,75)(75,0)
\psline[linewidth=0.75,arrowlength=1.5,arrowinset=0]{<-}(60,140)(60,92.5)
\pspolygon[linewidth=0.75](7.5,65)(52.5,65)(52.5,47.5)(7.5,47.5)
\pspolygon[linewidth=0.75](37.5,92.5)(82.5,92.5)(82.5,75)(37.5,75)
\psline[linewidth=0.75,arrowlength=1.5,arrowinset=0]{<-}(45,75)(45,65)
\rput[r](73.75,10){$\inputttt$}
\rput(60,83.75){\rref}
\rput[l](61.88,130){\outputtt}
\rput[r](43.75,11.25){\inputt}
\rput(30,56.88){\xxam}
\rput[l](46.88,70){$\recordd$}
\end{pspicture}
\end{center}
\end{minipage}
\caption{String diagrams: causation flows upwards}
\label{Fig:strings}
\end{center}
\end{figure}
Henceforth, we zoom out, hide the "innards" of causal models, and study how they are composed and decomposed. Towards this goal, {\causations} are presented as  \emph{string diagrams}, like in Fig.~\ref{Fig:strings}. String diagrams consist of just two graphic components:
\begin{itemize}
\item \textbf{\emph{strings}} --- representing \emph{types}, and
\item \textbf{\emph{boxes}} --- representing causal \emph{processes}.
\end{itemize}
Formally, types and processes are the basic building blocks of a causal modeling language: e.g., they may correspond to random variables and stochastic processes. Informally, a type can be thought of as a collection of events. In a causal process, enclosed in a box, the events of the input type, corresponding to the string that enters the box at the bottom, cause the events of the output type, corresponding to the string that exits the box at the top. The time just flows upwards in our string diagrams. We call the types consumed and produced by a causal process inputs and outputs (and not causes and effects) to avoid the ambiguities arising from the fact that the events produced by one process as effects may be consumed by another process as causes. The input and the output type may be the same. The diagram in Fig.~\ref{Fig:strings} on the left is hoped to convey an idea what is what. Presenting {\causations} as boxes allows us to abstract away the details when we need a high-level view, and also to refine the view as needed, by opening some boxes and displaying more details. This is similar to the mechanism of  virtual functions in programming: we specify the input and the output types, but postpone the implementation details.  A refined view of a {\causation} in an open box may be composed of other boxes connected by strings. An example\footnote{This has been a running example in causal modeling textbooks and lectures at least since \cite{koller2009probabilistic}.} is in Fig.~\ref{Fig:strings} on the right.  The types of the strings coming in and out show that \emph{grades} and \emph{reference letters}, as events produced as effects of a {\causation}  \textbf{course}, are causally determined by \emph{students' work} and \emph{class activities}, as well as by the \emph{hardness}\/ of the course itself. All these causal factors are viewed of events of suitable types. When we open the \textbf{course} box, we see that this {\causation} is composed from two simpler {\causations}: one is \textbf{exam and homework}, the other \textbf{reference request}. Each of them inputs two causal factors at the bottom; \textbf{exam and homework} outputs two effects, whereas \textbf{reference request} outputs one. Each  \emph{grade}, as an event of type Grade Report, is caused both by the \emph{course hardness} and by  the \emph{student work}; whereas the \emph{reference letters} are caused (influenced, determined\ldots) by the \emph{class activity} and by the course \emph{record}, which is again an effect of the \emph{student work}\/ and of the \emph{course hardness} in the process of \textbf{exam and homework}.  The causal dependencies between the random variables corresponding to the types Grade Report and Reference Letter, and their dependencies on the random variables corresponding to Course Hardness, Student Work, and Class Activity, are thus displayed in the string diagram inside the \textbf{course} box.  For a still more detailed view of causal relations, we could zoom in further, and open the boxes \textbf{exam and homework} and \textbf{reference request}, to display the dependencies through some other random variables, influenced by some other {\causations}.

The causes corresponding to the input types are assumed to be independent on each other. More precisely, the random variables, corresponding to the types Course Hardness, Student Work, and Class Activity, are assumed to be statistically independent. On the other hand, the causal dependencies of random variables corresponding to Grade Report, Record, and Reference Letter are displayed in the diagram. E.g., the content of a Reference Letter is not directly caused by Course Hardness, but it indirectly depends on it, since the student performance in the Record depends on it, and the Record is taken into account in the Reference Letter.

The abstract view of the {\causation} \textbf{box match} on the left could be refined in a similar way. The causes are boxers' actions, the effects are boxers' states. The images display a particular cause, and a particular effect of these types. The direct cause of the particular effect that is displayed on top is a particular blow, that is only visible inside the box. The cause at the bottom is the boxer's decision to deliver that particular blow. The {\causation} transforming this causal decision into its effect can be refined all the way to distinguishing good boxing from bad boxing, and to analyzing the causes of winning or losing. 

Spelling out a process corresponding to the \textbf{box match}  on the right  in Fig.~\ref{Fig:ali} might be even more amusing. While the strings outside the box would be the same, the causal dependencies inside the box would be different, as boxers' states are not caused by their blows, but feigned; and the blows are not caused by boxers' decisions, but by the movie director's requests.

\subsection{A category of {\causations}}\label{Sec:category}
String diagrams described in the previous section provide a graphic interface for any of causal modeling languages, irrespective of their details, displaying only how {\causations} are composed. It is often convenient to arrange compositional structures into \emph{categories} \cite{MacLaneS:CWM}. The categorical structures capturing causations turn out to yield to string diagrams \cite{Coecke-Spekkens:Bayes,CoeckeB:book,JacobsB:causal}. 
For the most popular models, as mentioned above, the strings correspond to random variables, the boxes to parametrized stochastic processes. The strings are thus the objects, the boxes the morphisms of a monoidal category. When no further constraints are imposed, this category can be viewed as the coslice, taken from the one-point space of the category of free algebras for a suitable stochastic monad. For a categorical account of full stochastic calculus, the monad can be taken over  measurable spaces 
\cite{Culbertson-Sturtz,GiryM:monad}. For a simplified account the essential structures, the convexity monad over sets may suffice \cite{JacobsB:causal}. For our purpose of capturing causal explanations \emph{and}\/ taking them into account as causal factors, we must deviate from the extant frameworks \cite{Coecke-Spekkens:Bayes,Culbertson-Sturtz,GiryM:monad,JacobsB:causal}\footnote{There is an entire research area of categorical probability theory, steadily built up since at least the 1960s, with hundreds of references. The three cited papers are a tiny random sample, which an interested reader could use as a springboard into further references, in many different flavors.} in two ways. One is minor, and well tried out: the randomness is captured by \emph{sub}\/probability measures, like in \cite{PanangadenP:book}, to account for nontermination. The other is that {\causations} must be taken up to \emph{stochastic indistinguishability}, for the reasons explained in \cite[Ch.~4]{spirtes2000causation}. 

\paragraph{Notation.} In the rest of the paper, we fix an abstract universe $\CCC$ of {\causations}, which happens to be a strict monoidal category \cite[Sec.~VII.1]{MacLaneS:CWM}. This structure is precisely what the string diagrams display. The categorical notation collects all strings together, in the family of event \emph{types}  $|\CCC|$; and for any pair of types $A,B \in |\CCC|$ it collects all boxes together, in the family $\CCC(A,B)$ of causal \emph{processes}. In summary,
\begin{itemize}
\item \textbf{\emph{strings}} form in the family of \emph{types} 
\bear |\CCC| & = & \{A, B, C\ldots, \mbox{Student Work}, \mbox{Grade Report}\ldots\}
\eear  
\item\textbf{\emph{boxes}} form for each pair $A,B\in |\CCC|$ the family of \emph{processes}, 
\bear \CCC(A,B) & = & \{f, g, t, u,\ldots, \mbox{\bf box match}, \mbox{\bf course}\ldots\}
\eear 
\end{itemize}
The compositional structure is briefly described in the Appendix.

\section{Modeling: Causal models as  causal factors}\label{Sec:Expl}

\subsection{Modeling causal models} 
The experimenter has been recognized as a causal factor in experiments since the onset of science. Already Galileo's concept of causality, mentioned in Sec.~\ref{Sec:Causality}, requires that the experimenter manipulates a cause in order to establish how the effect depends on it. In modern theories of causation, this idea is refined to the concept of \emph{intervention}\/ \cite{HagmayerY:intervention}. In quantum mechanics, the effects of measurements  depend not only on the causal freedom of the measured particles, but also on   experimenters' own causal freedom \cite{BohrN:37,conway2009strong}. As mentioned in Sec.~\ref{Sec:Intro}, similar questions arise even in magic: if the magician manipulates the effects, what causes magician's manipulations?

But if the magician and the experimenter are {\causations} themselves, then they also occur, as boxes among boxes, in the universe $\CCC$ of {\causations}. And if the experimenter's causal hypotheses are caused by some causal factors themselves, then the universe $\CCC$ contains, among other types, a distinguished type of causal models $\DP$, where the experimenter outputs his hypotheses. For instance, $\DP$ can be the type of our string diagram models, like in Fig.~\ref{Fig:strings}. Or $\DP$ can be the type of (suitably restricted) stochastic processes; or of Bayesian nets, including the running example \cite[Fig.~3.3]{koller2009probabilistic}, on which Fig.~\ref{Fig:strings} is based.\footnote{One of the coauthors of this paper thinks of $\DP$ as typing the well-formed expressions in any of the suitable modeling languages, discussed in Sec.~\ref{Sec:models}. The other coauthor is inclined to include a wider gamut of causal theories, including those touched upon in Sec.~\ref{Sec:history}. Different theories may not only describe different models, but also prescribe different interpretations.}   
There are many different languages for describing different causations, and thus many different ways to think of the type $\DP$. Psychologist's descriptions of causal cognition differ from physicist's view of the {\causation} of experimentation. What is their common denominator? \emph{What distinguishes $\DP$ from other types?}

In the Sec.~\ref{Sec:Axioms}, we attempt to answer this question by specifying a structure, using three simple axioms, which should be carried by the type $\DP$ in all frameworks for causal modeling. The other way around, we take this structure to  mean that a framework where it  occurs describes causal modeling; that the blind men talking about different parts of an elephant are talking about the same animal. This is where the \textbf{assumption} stated at the end of Sec.~\ref{Sec:Causality} is used. 

But stating and using the axioms requires two basic syntactic features, which are explained next.

\subsection{Parametrizing and steering models and processes} \label{Sec:Params}
Suppose that we want to explore how learning environments influence the {\causation} \textbf{course} from Fig.~\ref{Fig:strings}. How does the causal impact of Course Hardness change depending on the \emph{school}\/ where a \textbf{course} takes place?  Abbreviating for convenience the cause and the effect types of the {\causation} \textbf{course} to
\bear
\Causess & = & {\rm Course\  Hardness\ttimes  Student\  Work \ttimes Class\  Activity}\\
\Effectss & = & {\rm Grade\  Report \ttimes Reference\  Letter}
\eear
we now have 
\begin{itemize}
\item a family of \causations\ $\Causess\  \tto{\ {\rm \bf course}\, (shool)\ } \  \Effectss$, indexed over\\  $school \in$  Schools, 
or equivalently
\item a parametric {\causation} $
{\rm Schools}\ttimes \Causess \ \tto{{\rm \bf course}} \  \Effectss$.
\end{itemize}
In the first case, each \causation\ \textbf{course}\emph{(school)} consumes the causal inputs in the form \textbf{course}\emph{(school)(hardness, work, activity)}; whereas in the second case, all inputs are consumed together, in the form  \textbf{course}\emph{(school, hardness, work, activity)}. The difference between the parameter \emph{school}\/ and the general causal factors $hardness, work, activity$ is that the parameter is \emph{deterministic}, whereas the general causal factors influence processes with some \emph{uncertainty}. This means that entering the same parameter $school$ always produces the same {\causation}, whereas the same causal factors $hardness, work, activity$ may have different effects in different samples.

The convenience of internalizing the indices and writing indexed families as parametric processes is that steering processes can be internalized as reparametrizing. E.g. given a function $\varsigma: {\rm Teachers} \to {\rm Schools}$, mapping each \emph{teacher}\/ the the \emph{school}\/ where they work, induces the reindexing of families
\[\prooftree
\Causess\  \tto{\ {\rm \bf course}\, (shool)\ } \  \Effectss\hspace{3em} school \in {\rm Schools}
\justifies
\Causess\  \tto{\ {\rm \bf course}\, \left(\varsigma\, (teacher)\right)\ } \  \Effectss\hspace{2em} teacher \in {\rm Teachers}
\endprooftree
\]
which can however be captured as a single causal process in the universe $\CCC$
\[
{\rm Teachers}\ttimes \Causess \tto{\  \varsigma\ \ttimes \ \Causess\ } {\rm Schools}\ttimes \Causess \ \tto{\ {\rm \bf course}\ } \  \Effectss
\]
where the causal factor $s$ happens to be deterministic.
%
%
In general, an arbitrary {\causation} $\Params \ttimes A \tto{\ {\rm \bf process}\ }  B$ can be steered along an arbitrary function ${\tt steer}:\Paramss \to \Params$, viewed as a deterministic process:
\[
\Paramss \ttimes A \tto{\ \tt steer\,  \ttimes\,  A\ } \Params \ttimes A \tto{\ {\rm \bf process}\ }  B\]
%
Deterministic functional dependencies are characterized in string diagrams in  Appendix~\ref{Sec:functions}.

\subsection{Axioms of causal cognition}\label{Sec:Axioms}
Every framework for causal modeling must satisfy the followin axioms:

\begin{enumerate}[label = {\bf \Roman* :}] 
\item Every causal model models a unique {\causation} (over the same parameters).

\item  Every {\causation} has a model (not necessarily unique).

\item Models are preserved under steering.
%
%
%
%
%
\end{enumerate}

\subsubsection{Axioms formally}
We view a causal model as a family $\xplan(y) \in \DP$, indexed by $y\in \Params$, i.e. as a parametrized model $\Params \tto{\ \xplan\ } \DP$. The notation introduced in Sec.~\ref{Sec:category} collects all 
$\Params$-parametrized causal models in the set of processes $\CCC(\Params, \DP)$. The parameter type $\Params$ is arbitrary. On the other hand, all $\Params$-parametrized causal processes $\Params \ttimes A \tto p B$, where the events of type $A$ cause events of type $B$, are collected in the set $\CCC(\Params \ttimes A, B)$. 

\paragraph{Axiom I} postulates that every parametrized causal model $\Params\tto{\xplan} \DP$ induces a unique causal process $\Params\ttimes A\tto{\Semantics{\xplan}} B$ with the same parameters $\Params$. A  causal modeling framework is thus given by a family of the \emph{\explains}\/ maps 
\bea\label{eq:Semantics}
\CCC(\Params, \DP) & \tto{\Semantics{-}} & \CCC(\Params \ttimes A, B)
\eea
indexed over all types $\Params, A$ and $B$. 

\paragraph{Axiom II} says that the {\explains} maps $\Semantics{-}$ are surjective: for every causal process $\Params\ttimes A\tto p B$ there is a causal model $\Params\tto P \DP$ that predicts its behavior, in the sense that the process $\Semantics{P}$ is \emph{indistinguishable} from $p$, which we write $\Semantics{P}\approx p$.  The indistinguishability relation $\approx$ is explained in the next section.

\paragraph{Axiom III} says that for any function $\varsigma: \Paramss\to \Params$ and any causal model $\Params\tto\xplan \DP$ reparametrizing $\xplan$ along $\varsigma$ models steering the modeled process $\Semantics \xplan$ along it, in the sense that
\bea\label{eq:III}
\Semantics{\Paramss\tto{\varsigma} \Params\tto\xplan \DP} & = & \Paramss\ttimes A \tto{\varsigma \ttimes A} \Params\ttimes A\tto{\Semantics{\xplan}} B
\eea

\subsubsection{Axioms informally}
\paragraph{Axiom I } is a soundness requirement: it says that every causal model models some causal process. If causal processes are viewed as observations, the axiom thus says that for any causal model, we will recognize the modeled process if we observe it.

\paragraph{Axiom II} is a completeness requirement: it says that  for  every {\causation} that we may observe, there is a causal model that models it. Can we really model everything that we observe? Yes, but our models are valid, i.e. their predictions are consistent with the behavior of the modeled processes \emph{only in so far as our current observations go}. Given a process $p$, we can always find a model $\xplan$ whose predictions $\Semantics{\xplan}$ summarize our observations of $p$, so that  $\Semantics{\xplan}$ and $p$ are for us \emph{indistinguishable}, i.e. $\Semantics{P}\approx p$. The less we observe, the less we distinguish, the easier we model. 

\paragraph{Indistinguishability relations}, be it statistical, computational, or observational, are central in experimental design, in theory of computation, and in modern theories of causation \cite[Ch.~4]{spirtes2000causation}. The problem of distinguishing between observed processes has been tackled since the early days of statistics by significance testing, and since the early days of computation by various semantical and testing equivalences. Axiom II says that, up to an indistinguishability relation, any observed causal process has a model. 
It does not say anything about the hardness of modeling. This problem is tackled in research towards \emph{cause discovery algorithms} \cite{spirtes2000causation}.


%
%
%

\paragraph{Axiom III} is a coherence requirement: it says that steering a process $\Params\ttimes A\tto p B$ along a deterministic function 
$\varsigma:\Paramss\to \Params$ does not change the causation, in the sense of \eqref{eq:III} or  
\bea\label{eq:steering}
\Semantics P \approx p &\implies & \Semantics{P\circ \varsigma} \approx \big(p\circ (\varsigma\ttimes A)\big)
\eea


\subsection{Universal testing}\label{Sec:universal}
Since any causal modeling language can thus be viewed as a type $\DP$, living in the process universe $\CCC$, the family of all causal models $\omega \in \DP$, trivially indexed over itself, can be represented by the identity function $\DP\tto{Id} \DP$, viewed as a $\DP$-parametrized model.   
 Instantiating in  \eqref{eq:III} $Id$ for $P$ (and thus $\DP$ for $\Params$) yields $\uev \varsigma = \uev{Id}\circ(\varsigma \ttimes A)$, for any $\Paramss\tto\varsigma \DP$. 
\begin{figure}[ht]
\begin{center}
\newcommand{\plann}{\begin{minipage}{1.5cm}\bf \scriptsize causal\\
\bf process\\ \centering $\displaystyle p$\end{minipage}}
\newcommand{\universal}{\begin{minipage}{1.5cm}\bf \scriptsize universal\\
\bf testing\\
\centering $\Semantics{Id}$\end{minipage}
}
\newcommand{\lab}{\includegraphics[height=2.7cm,width=1.8cm]{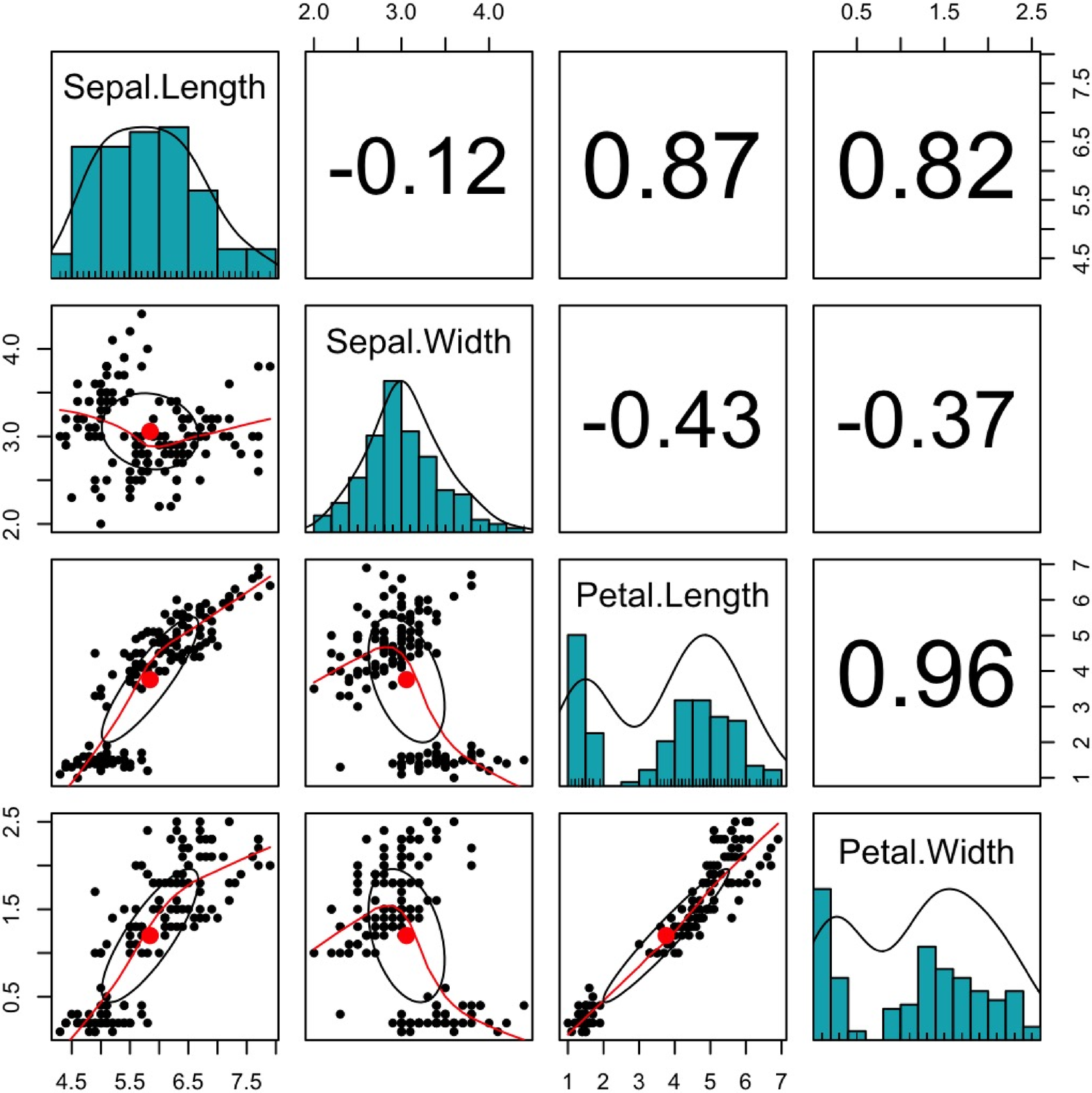}}
\newcommand{\outputtt}{$\scriptstyle B$}
\newcommand{\inputt}{$\scriptstyle A$}
\newcommand{\inputtt}{\scriptstyle \Paramss}
\newcommand{\Language}{$\scriptscriptstyle \DP$}
\newcommand{\graph}{\includegraphics[height=1.8cm]{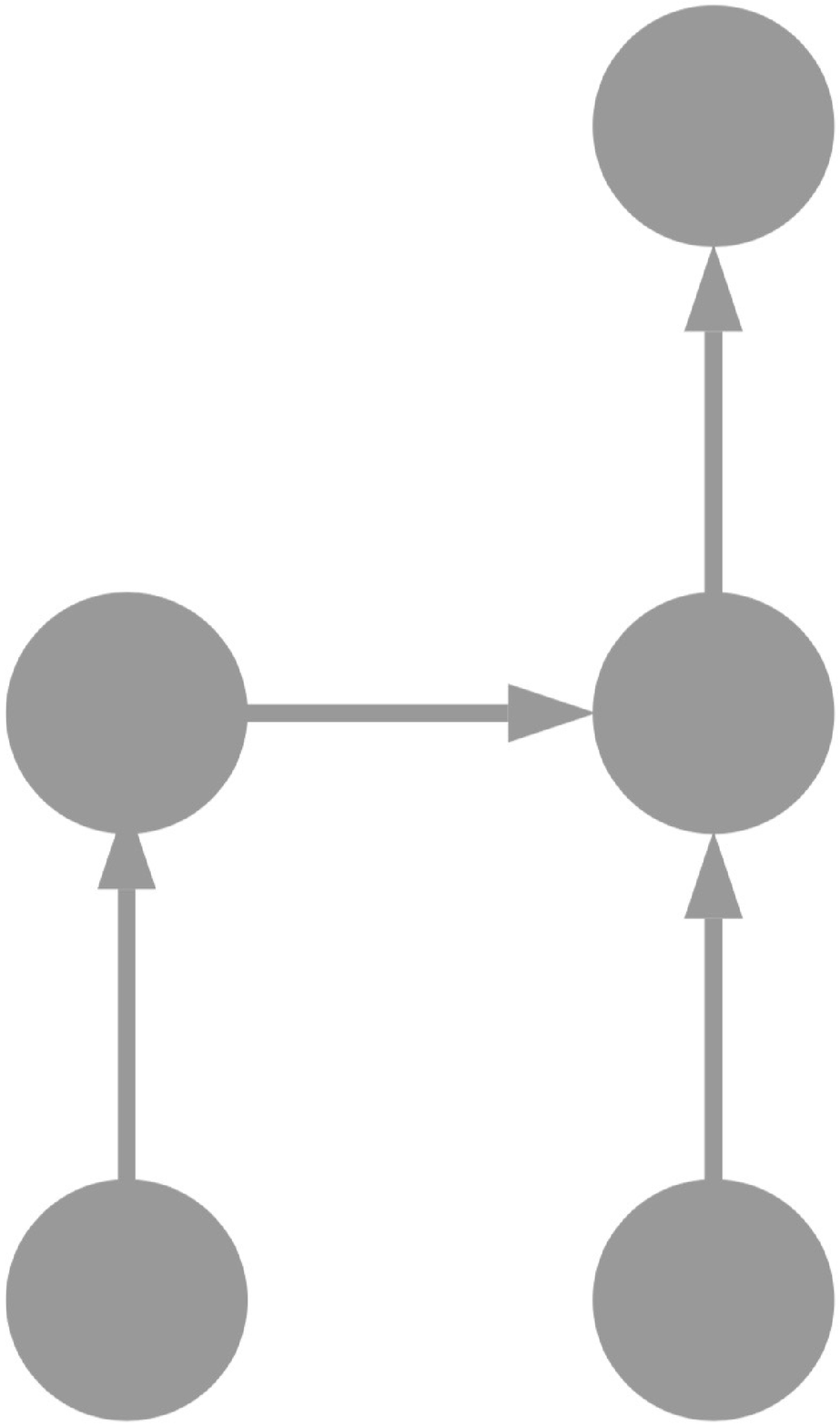}}
\newcommand{\program}{
\begin{minipage}{1cm}\bf \scriptsize model
\\[-.3ex]
\centering$\displaystyle P$
\end{minipage}}
\newcommand{\EQLS}{
\huge$\mathbf{\approx}$}
\def\JPicScale{.6}
\ifx\JPicScale\undefined\def\JPicScale{1}\fi
\psset{unit=\JPicScale mm}
\psset{linewidth=0.3,dotsep=1,hatchwidth=0.3,hatchsep=1.5,shadowsize=1,dimen=middle}
\psset{dotsize=0.7 2.5,dotscale=1 1,fillcolor=black}
\psset{arrowsize=1 2,arrowlength=1,arrowinset=0.25,tbarsize=0.7 5,bracketlength=0.15,rbracketlength=0.15}
\begin{pspicture}(0,0)(240,122.5)
\psline[linewidth=0.75](15,50)(15,0)
\rput[r](12.5,5){$\inputtt$}
\psline[linewidth=0.75](75,50)(75,0)
\psline[linewidth=0.75,arrowlength=1.5,arrowinset=0]{<-}(65,122.5)(65,95)
\newrgbcolor{userLineColour}{0.6 0.6 0.6}
\pspolygon[linewidth=0.75,linecolor=userLineColour](5,70)(55,70)(55,55)(5,55)
\newrgbcolor{userLineColour}{0.6 0.6 0.6}
\pspolygon[linewidth=0.75,linecolor=userLineColour](35,93.75)(85,93.75)(85,80)(35,80)
\newrgbcolor{userLineColour}{0.6 0.6 0.6}
\psline[linewidth=0.75,linecolor=userLineColour,arrowlength=1.5,arrowinset=0]{<-}(45,80)(45,70)
\rput[l](68.12,114.38){\outputtt}
\rput[r](72.5,5){\inputt}
\psline[linewidth=0.75](165,100)(240,100)
\psline[linewidth=0.75](150,70)(150,20)
\psline[linewidth=0.75](215,50)(240,50)
\psline[linewidth=0.75](150,20)(215,20)
\rput[br](198.75,25){\program}
\rput(122.5,70){\EQLS}
\psline[linewidth=0.75,arrowlength=1.5,arrowinset=0]{<-}(165,20)(165,0)
\rput[r](162.5,5){$\inputtt$}
\psline[linewidth=0.75,arrowlength=1.5,arrowinset=0]{<-}(227.5,50)(227.5,0)
\psline[linewidth=0.75,arrowlength=1.5,arrowinset=0]{<-}(227.5,122.5)(227.5,100)
\rput[l](230.62,115){\outputtt}
\rput[r](225,5){\inputt}
\newrgbcolor{userLineColour}{0.6 0.6 0.6}
\psline[linewidth=0.75,linecolor=userLineColour,arrowlength=1.5,arrowinset=0]{<-}(15,55)(15,50)
\newrgbcolor{userLineColour}{0.6 0.6 0.6}
\psline[linewidth=0.75,linecolor=userLineColour,arrowlength=1.5,arrowinset=0]{<-}(75,80)(75,50)
\newrgbcolor{userLineColour}{0.6 0.6 0.6}
\psline[linewidth=0.75,linecolor=userLineColour](65,100)(65,93.75)
\pspolygon[linewidth=0.75](0,100)(90,100)(90,50)(0,50)
\rput(163.75,40){\graph}
\psline[linewidth=0.75](165,70)(215,20)
\psline[linewidth=0.75,arrowlength=1.5,arrowinset=0]{<-}(190,75)(190,45)
\rput[tl](5,95){$\plann$}
\rput(225,75){\lab}
\psline[linewidth=0.75](165,100)(215,50)
\rput[tl](185,95){\universal}
\psline[linewidth=0.75](240,50)(240,100)
\psline[linewidth=0.75](150,70)(165,70)
\rput(0,0){\rput(190,60){\psframebox[fillcolor=white,fillstyle=solid]{\Language}}}
\end{pspicture}
\caption{There is an explanation for every causation}
\label{Fig:interpreter}
\end{center}
\end{figure}
\emph{Mutatis mutandis}, for an arbitrary process $p$ and a model $P$ assured by axioms I and II, axiom II thus implies $p \approx \Semantics{Id}\circ(P \ttimes A)$, as displayed in Fig.~\ref{Fig:interpreter}. 

Any causal universe $\CCC$, as soon as it satisfies axioms I--III, thus contains, for any pair $A,B$, a \textbf{\emph{universal testing}}\/ process $\DP\ttimes A\tto{\Semantics{Id}} B$, which inputs causal models and tests their predictions, in the sense that $\Paramss\ttimes A \tto{P\otimes A}\DP\ttimes A \tto{\Semantics{Id}} B$ is indistinguishable from $\Paramss \ttimes A\tto p B$ whenever $\Semantics{P}$ is indistinguishable from $p$. Universal testing is thus a causal process where the predictions of causal models are derived as their effects. It can thus be construed as a very high level view of scientific practice; perhaps also as an aspect of cognition.

\subsection{Partial modeling}\label{Sec:partial}
If a causal process has multiple causal factors, then they can be modeled separately be treating some of them as model parameters. E.g. a process in the form $\Params\ttimes \Paramss \ttimes A \tto r B$ can be viewed as a $\Params$-parametrized process with causal factors of type $\Paramss\ttimes A$, or as a $\Params\ttimes \Paramss$-parametrized process with causal factors of type $A$. The two different instances of \eqref{eq:Semantics}, both surjections by axiom II, would lead to models $\Params \tto{R'} \DP$, and $\Params\ttimes \Paramss \tto{R''} \DP$, with $r\approx \Semantics{R'}_\Params \approx \Semantics{R''}_{\Params\ttimes\Paramss}$ by \eqref{eq:Semantics}, and thus $r\approx \Semantics{Id}_{\Paramss\ttimes A}\circ (R'\ttimes \Paramss\ttimes A) \approx \Semantics{Id}_{A}\circ (R'' \ttimes A)$ by Fig.~\ref{Fig:interpreter}.
\begin{figure}[ht]
\begin{center}
\newcommand{\plann}{\begin{minipage}{2cm}\bf \footnotesize universal\\
\bf testing\\
\centering $\Semantics{Id}_{\Paramss\ttimes A}$\end{minipage}}
\newcommand{\lab}{\begin{minipage}{2cm}\bf \footnotesize universal\\
\bf testing\\
\centering $\Semantics{Id}_A$\end{minipage}
}
\newcommand{\outputtt}{\mbox{\scriptsize $B$}}
\newcommand{\inputt}{\mbox{\scriptsize $A$}}
\newcommand{\inputs}{\mbox{\scriptsize $\Paramss$}}
\newcommand{\inputtt}{%
\mbox{\scriptsize $\Paramss$}
}
\newcommand{\Language}{\tiny $\DP$}
\newcommand{\Biglanguage}{$\scriptstyle \DP$}
\newcommand{\pprogram}{$\Xi$}
\newcommand{\program}{{\scriptsize \tt model}}
\newcommand{\EQLS}{
\huge$\mathbf{\approx}$}
\def\JPicScale{.6}
\ifx\JPicScale\undefined\def\JPicScale{1}\fi
\psset{unit=\JPicScale mm}
\psset{linewidth=0.3,dotsep=1,hatchwidth=0.3,hatchsep=1.5,shadowsize=1,dimen=middle}
\psset{dotsize=0.7 2.5,dotscale=1 1,fillcolor=black}
\psset{arrowsize=1 2,arrowlength=1,arrowinset=0.25,tbarsize=0.7 5,bracketlength=0.15,rbracketlength=0.15}
\begin{pspicture}(0,0)(240,112.5)
\psline[linewidth=0.75](170,90)(240,90)
\psline[linewidth=0.75](210,50)(240,50)
\psline[linewidth=0.75](180,20)(210,20)
\rput(125,70){\EQLS}
\psline[linewidth=0.75,arrowlength=1.5,arrowinset=0]{<-}(192.5,20)(192.5,0)
\rput[r](191.25,5){$\inputtt$}
\psline[linewidth=0.75,arrowlength=1.5,arrowinset=0]{<-}(225,50)(225,0)
\psline[linewidth=0.75,arrowlength=1.5,arrowinset=0]{<-}(225,112.5)(225,90)
\rput[l](226.25,103.75){\outputtt}
\rput[r](223.75,5){\inputt}
\psline[linewidth=0.75](170,60)(210,20)
\psline[linewidth=0.75,arrowlength=1.5,arrowinset=0]{<-}(190,70)(190,40)
\rput(220,70){\lab}
\psline[linewidth=0.75](170,90)(210,50)
\psline[linewidth=0.75](240,50)(240,90)
\psline[linewidth=0.75](140,60)(170,60)
\rput(0,0){\rput(190.62,53.12){\psframebox[fillcolor=white,fillstyle=solid]{\Language}}}
\psline[linewidth=0.75](140,60)(180,20)
\psline[linewidth=0.75,arrowlength=1.5,arrowinset=0]{<-}(160,40)(160,0)
\rput[r](158.75,5){\Biglanguage}
\psline[linewidth=0.75](0,90)(100,90)
\psline[linewidth=0.75](40,50)(100,50)
\psline[linewidth=0.75,arrowlength=1.5,arrowinset=0]{<-}(55,50)(55,0)
\psline[linewidth=0.75,arrowlength=1.5,arrowinset=0]{<-}(85,50)(85,0)
\psline[linewidth=0.75,arrowlength=1.5,arrowinset=0]{<-}(85,112.5)(85,90)
\rput[l](86.25,103.75){\outputtt}
\rput[r](83.75,5){\inputt}
\psline[linewidth=0.75](0,90)(40,50)
\psline[linewidth=0.75](100,50)(100,90)
\psline[linewidth=0.75,arrowlength=1.5,arrowinset=0]{<-}(20,70)(20,0)
\rput[r](18.75,5){\Biglanguage}
\rput[r](53.75,5){\inputs}
\rput(175,40){\pprogram}
\rput(65,70){\plann}
\end{pspicture}
\caption{Causes of type $\Paramss$ are treated as parameters}
\label{Fig:specializer}
\end{center}
\end{figure}
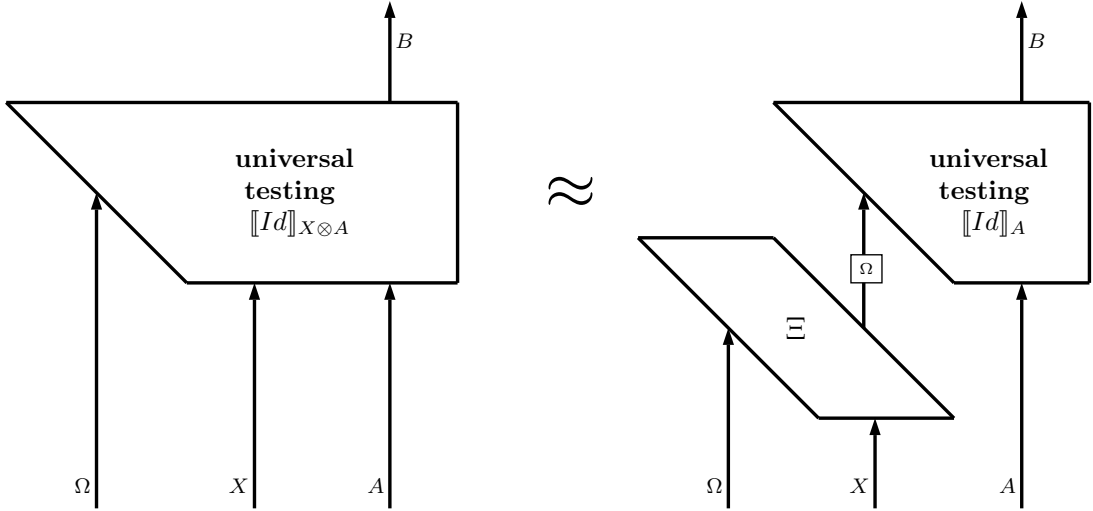
 
In particular, taking $r$ to be the universal process $\DP \ttimes \Paramss \ttimes A \tto{\Semantics{Id}} B$, and interpreting it as an $\DP\ttimes \Paramss$-parametrized process, leads to an $\DP\ttimes \Paramss$-parametrized model $\Xi$, such that $\Semantics{Id}_{\Paramss\ttimes A} \approx \Semantics{Id}_{A}\circ (\Xi \ttimes A)$, as displayed in Fig.~\ref{Fig:specializer}.

\subsection{Slicing models}
Using universal testing processes and partial modeling, causal processes can be modeled incrementally, factor by factor, like in Fig.~\ref{Fig:slicing}.
\newcommand{\program}{\tiny \tt model}
\newcommand{\progpart}{\tiny \tt model}
\newcommand{\gee}{\begin{minipage}{1.5cm}\bf \footnotesize causal\\
\bf \footnotesize process\centering\end{minipage}}
\newcommand{\hee}{\bf \footnotesize testing}
\newcommand{\eqlsa}{\stackrel {(a)}\approx
}
\newcommand{\eqlsb}{\stackrel {(b)}\approx
}
\newcommand{\eqlsc}{\stackrel {(c)}\approx
}
\newcommand{\pprogram}{{\tiny \tt partial}}
\newcommand{\ppprogram}{\begin{minipage}{1.5cm}\tt \scriptsize steered\\
\tt \centering model\end{minipage}
}
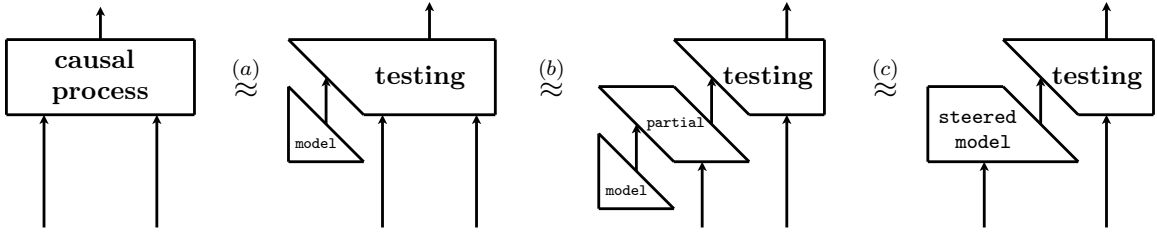
\begin{figure}[htbp]
\begin{center}
\def\JPicScale{.5}
\ifx\JPicScale\undefined\def\JPicScale{1}\fi
\psset{unit=\JPicScale mm}
\psset{linewidth=0.3,dotsep=1,hatchwidth=0.3,hatchsep=1.5,shadowsize=1,dimen=middle}
\psset{dotsize=0.7 2.5,dotscale=1 1,fillcolor=black}
\psset{arrowsize=1 2,arrowlength=1,arrowinset=0.25,tbarsize=0.7 5,bracketlength=0.15,rbracketlength=0.15}
\begin{pspicture}(0,0)(305,60)
\psline[linewidth=0.75]{<-}(100,30)(100,0)
\psline[linewidth=0.75]{<-}(125,30)(125,0)
\psline[linewidth=0.75](95,30)(75,50)
\psline[linewidth=0.75](75,50)(130,50)
\psline[linewidth=0.75](95,30)(130,30)
\psline[linewidth=0.75](130,50)(130,30)
\psline[linewidth=0.75]{<-}(112.5,60)(112.5,50)
\psline[linewidth=0.75]{<-}(85,40)(85,27.5)
\psline[linewidth=0.75]{<-}(167.5,27.5)(167.5,15)
\psline[linewidth=0.75]{<-}(185,17.5)(185,0)
\psline[linewidth=0.75](157.5,37.5)(177.5,37.5)
\psline[linewidth=0.75](197.5,17.5)(177.5,17.5)
\psline[linewidth=0.75](197.5,30)(177.5,50)
\psline[linewidth=0.75](177.5,37.5)(197.5,17.5)
\psline[linewidth=0.75](177.5,50)(217.5,50)
\psline[linewidth=0.75]{<-}(187.5,40)(187.5,27.5)
\psline[linewidth=0.75](197.5,30)(217.5,30)
\psline[linewidth=0.75](217.5,50)(217.5,30)
\psline[linewidth=0.75](157.5,37.5)(177.5,17.5)
\psline[linewidth=0.75]{<-}(207.5,60)(207.5,50)
\psline[linewidth=0.75]{<-}(207.5,30)(207.5,0)
\psline[linewidth=0.75](157.5,25)(157.5,5)
\psline[linewidth=0.75](157.5,5)(177.5,5)
\rput(165,10){\program}
\psline[linewidth=0.75](157.5,25)(177.5,5)
\psline[linewidth=0.75](75,17.5)(95,17.5)
\rput(82.5,22.5){\program}
\psline[linewidth=0.75](75,37.5)(95,17.5)
\psline[linewidth=0.75](75,37.5)(75,17.5)
\psline[linewidth=0.75]{<-}(40,30)(40,0)
\psline[linewidth=0.75](0,30)(0,50)
\psline[linewidth=0.75](0,50)(50,50)
\psline[linewidth=0.75](0,30)(50,30)
\psline[linewidth=0.75](50,50)(50,30)
\psline[linewidth=0.75]{<-}(25,58.75)(25,50)
\rput(25,40){\gee}
\psline[linewidth=0.75]{<-}(10,30)(10,0)
\psline[linewidth=0.75]{<-}(292.5,30)(292.5,0)
\psline[linewidth=0.75](245,17.5)(245,37.5)
\psline[linewidth=0.75](265,50)(305,50)
\psline[linewidth=0.75](285,30)(305,30)
\psline[linewidth=0.75](305,50)(305,30)
\psline[linewidth=0.75]{<-}(292.5,60)(292.5,50)
\psline[linewidth=0.75]{<-}(260,17.5)(260,0)
\psline[linewidth=0.75](285,30)(265,50)
\psline[linewidth=0.75](245,17.5)(285,17.5)
\psline[linewidth=0.75](245,37.5)(265,37.5)
\psline[linewidth=0.75](285,17.5)(265,37.5)
\psline[linewidth=0.75]{<-}(275,40)(275,27.5)
\rput(145,40){$\eqlsb$}
\rput(233.75,40){$\eqlsc$}
\rput(63.75,40){$\eqlsa$}
\rput(110,40){\hee}
\rput(290,40){\hee}
\rput(202.5,40){\hee}
\rput(178.12,27.5){\pprogram}
\rput(261.25,26.25){\ppprogram}
\end{pspicture}
\caption{Separating independent causes allows incremental modeling}
\label{Fig:slicing}
\end{center}
\end{figure}
%
%


\section{Construction: Self-confirming causal models}\label{Sec:Factors}

In everyday life, we often defend our views by steering ourselves to the standpoints where they seem valid. This assures belief perseverance, but also leads to cognitive bias and self-deception. In science, experimenters obviously influence the outcomes of their experiments by choosing what to measure. In quantum mechanics, this influence propagates to processes themselves, as the choice of the measurement basis determines the choice of the states to which the measured process may collapse \cite{Bohr51,conway2009strong}. Causal models are thus always among the causal factors of the modeled quantum processes\footnote{"The complete freedom of the procedure in experiments common to all investigations of physical phenomena, is in itself of course contained in our free choice of the experimental arrangement." \cite{BohrN:38}}. Formally, a quantum process is thus always in the form $\DP \ttimes A \tto q B$, parametrized by its models in $\DP$.

Many processes outside the realm of quantum are in this form as well, as causal  cognition curls into itself by influencing causation. The placebo and the nocebo effects are ubiquitous: patient's belief in the effectiveness of a medication contributes to its effectiveness, whereas negative beliefs often cause negative effects. In Shakespeare's tragedy, Macbeth is driven to murder the King by the prophecy that he would murder the King. In the early days of Facebook, attracting new members required convincing them that many of their friends were already members. This initially had to be a lie, but many believed it, joined, and it ceased to be a lie. 

Other causal models impact their own validity negatively. If the customers of a restaurant come to believe that it is too busy on Fridays, it may end up empty on Fridays: the belief will invalidate itself. If the rumor that it was empty spreads, the customers will swarm back, and the model will invalidate itself again. Such network effects are observed not only in the famous El Farol bar in Santa Fe, and in Kolkatta paise restaurants, but also in financial and stock markets, and in urban agglomerations. Causal models often influence the modeled causal structures, and sometimes impact their own validity. --- \emph{What are the conditions and limitations of this phenomenon?}

\begin{figure}[ht]
\newcommand{\Beliefs}{\tiny $\DP$}
\newcommand{\Causes}{$\scriptstyle A$}
\newcommand{\Effects}{$\scriptstyle B$}
\newcommand{\universal}{$\Semantics{Id}$}
\renewcommand{\gee}{$q$}
\renewcommand{\program}{$\Gamma$}
\newcommand{\EQLS}{\Large \mbox{$\approx$}}
\begin{center}
\def\JPicScale{.6}
\ifx\JPicScale\undefined\def\JPicScale{1}\fi
\psset{unit=\JPicScale mm}
\psset{linewidth=0.3,dotsep=1,hatchwidth=0.3,hatchsep=1.5,shadowsize=1,dimen=middle}
\psset{dotsize=0.7 2.5,dotscale=1 1,fillcolor=black}
\psset{arrowsize=1 2,arrowlength=1,arrowinset=0.25,tbarsize=0.7 5,bracketlength=0.15,rbracketlength=0.15}
\begin{pspicture}(0,0)(140,76.88)
\psline[linewidth=0.75](90,65)(140,65)
\psline[linewidth=0.75](90,30.62)(90,5.62)
\psline[linewidth=0.75](140,40)(140,65)
\psline[linewidth=0.75]{<-}(130,76.88)(130,65)
\psline[linewidth=0.75](115,40)(140,40)
\psline[linewidth=0.75]{<-}(130,40)(130,-0.62)
\psline[linewidth=0.75](90,5.62)(115,5.62)
\rput(122.5,52.5){\universal}
\rput(98.75,12.5){\program}
\rput(75,51.25){\EQLS}
\psline[linewidth=0.75](90,30.62)(115,5.62)
\psline[linewidth=0.75]{<-}(102.5,52.5)(102.5,18.75)
\psline[linewidth=0.75](90,65)(115,40)
\psline[linewidth=0.75](5,65)(55,65)
\psline[linewidth=0.75](2.5,30.62)(2.5,5.62)
\psline[linewidth=0.75](55,40)(55,65)
\psline[linewidth=0.75]{<-}(30,76.88)(30,65)
\psline[linewidth=0.75](5,40)(55,40)
\psline[linewidth=0.75]{<-}(45,40)(45,-0.62)
\psline[linewidth=0.75](2.5,5.62)(27.5,5.62)
\rput(11.25,12.5){\program}
\psline[linewidth=0.75](2.5,30.62)(27.5,5.62)
\psline[linewidth=0.75]{<-}(15,40)(15,18.75)
\psline[linewidth=0.75](5,65)(5,40)
\rput(30,52.5){$\gee$}
\rput[l](46.25,3.75){\Causes}
\rput(0,0){\rput(15,30){\psframebox[fillcolor=white,fillstyle=solid]{\Beliefs}}}
\rput[l](131.25,3.75){\Causes}
\rput(0,0){\rput(102.5,30.62){\psframebox[fillcolor=white,fillstyle=solid]{\Beliefs}}}
\rput[r](28.75,70.62){\Effects}
\rput[r](128.75,70.62){\Effects}
\end{pspicture}
\caption{The process $q$ steered by a self-confirming model $\Gamma$ into  $q(\Gamma, a) \approx \Semantics{\Gamma}(a)$}
\label{Fig:kleene}
\end{center}
\end{figure}
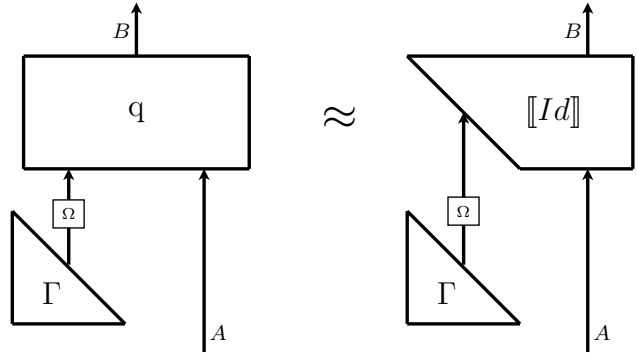

Fig.~\ref{Fig:kleene} formalizes this question. Under which conditions can a process 
$\DP \ttimes A \tto q B$ be steered  by a model $\Gamma \in \DP$ to a process which confirms $\Gamma$'s predictions, i.e. such that $q\circ(\Gamma\ttimes A) \approx \Semantics \Gamma$?
By reinterpreting the outputs of $q$, under certain additional conditions, the same schema can be used to construct models that would invalidate themselves. 

In any case, it turns out that a causal universe $\CCC$ \emph{contains a self-confirming causal model for every causal process parametrized over models}, as soon as it supports axioms I--III. 

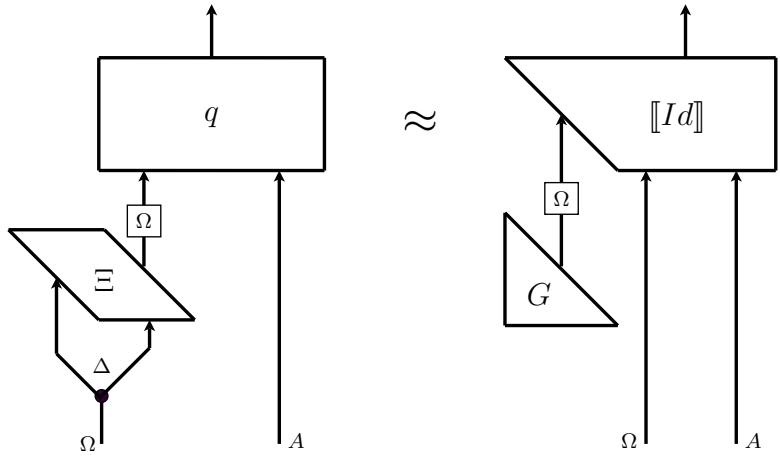
\begin{figure}[ht]
\begin{center}
\newcommand{\universal}{$\Semantics{Id}$}
\renewcommand{\gee}{q}
\renewcommand{\program}{$G$}
\newcommand{\EQLS}{\Large $\approx$}
\newcommand{\parteval}{\Xi}
\newcommand{\Branch}{$\scriptstyle \Delta$}
\newcommand{\Beliefs}{$\scriptstyle \DP$}
\newcommand{\Causes}{$\scriptstyle A$}
\def\JPicScale{.6}
\ifx\JPicScale\undefined\def\JPicScale{1}\fi
\psset{unit=\JPicScale mm}
\psset{linewidth=0.3,dotsep=1,hatchwidth=0.3,hatchsep=1.5,shadowsize=1,dimen=middle}
\psset{dotsize=0.7 2.5,dotscale=1 1,fillcolor=black}
\psset{arrowsize=1 2,arrowlength=1,arrowinset=0.25,tbarsize=0.7 5,bracketlength=0.15,rbracketlength=0.15}
\begin{pspicture}(0,0)(173.75,97.5)
\psline[linewidth=0.75](113.75,85.62)(173.75,85.62)
\psline[linewidth=0.75](113.75,51.25)(113.75,26.25)
\psline[linewidth=0.75](173.75,60.62)(173.75,85.62)
\psline[linewidth=0.75]{<-}(153.75,97.5)(153.75,85.62)
\psline[linewidth=0.75](138.75,60.62)(173.75,60.62)
\psline[linewidth=0.75]{<-}(165,60.62)(165,0)
\psline[linewidth=0.75](113.75,26.25)(138.75,26.25)
\rput(151.87,72.5){\universal}
\rput(121.25,33.12){\program}
\rput(95,71.25){\EQLS}
\psline[linewidth=0.75](113.75,51.25)(138.75,26.25)
\psline[linewidth=0.75]{<-}(126.25,73.13)(126.25,39.37)
\psline[linewidth=0.75](113.75,85.62)(138.75,60.62)
\psline[linewidth=0.75](23.75,85.63)(73.75,85.62)
\psline[linewidth=0.75](73.75,60.62)(73.75,85.62)
\psline[linewidth=0.75]{<-}(48.75,97.5)(48.75,85.62)
\psline[linewidth=0.75](23.75,60.63)(73.75,60.62)
\psline[linewidth=0.75]{<-}(63.75,60.62)(63.75,0)
\psline[linewidth=0.75]{<-}(33.75,60.63)(33.75,39.37)
\psline[linewidth=0.75](23.75,85.63)(23.75,60.63)
\rput(48.75,72.5){$\gee$}
\psline[linewidth=0.75](3.75,47.49)(25,47.49)
\psline[linewidth=0.75](45,27.5)(23.75,27.5)
\psline[linewidth=0.75](25,47.49)(45,27.5)
\psline[linewidth=0.75](3.75,47.49)(23.75,27.5)
\psline[linewidth=0.75](24.38,10.63)(24.38,0)
\newrgbcolor{userFillColour}{0.2 0 0.2}
\rput{0}(24.37,10.62){\psellipse[fillcolor=userFillColour,fillstyle=solid](0,0)(1.41,-1.41)}
\psline[linewidth=0.75]{<-}(35,27.5)(35,21.25)
\psline[linewidth=0.75]{<-}(14.38,36.88)(14.38,20.01)
\psline[linewidth=0.75](35,21.25)(24.38,10.63)
\psline[linewidth=0.75](14.38,20.01)(24.38,10)
\psline[linewidth=0.75]{<-}(145,60.62)(145,0)
\rput(25,36.25){$\parteval$}
\rput[r](143.12,0.62){\Beliefs}
\rput[l](166.88,0.62){\Causes}
\rput(0,0){\rput(126.25,54.38){\psframebox[fillcolor=white,fillstyle=solid]{\Beliefs}}}
\rput[r](23.12,0){\Beliefs}
\rput[l](65.62,0.62){\Causes}
\rput(0,0){\rput(33.75,49.38){\psframebox[fillcolor=white,fillstyle=solid]{\Beliefs}}}
\rput[b](24.38,15.62){\Branch}
\end{pspicture}
\caption{Modeling self-modeling}
\label{Fig:fixp-1}
\end{center}
\end{figure}

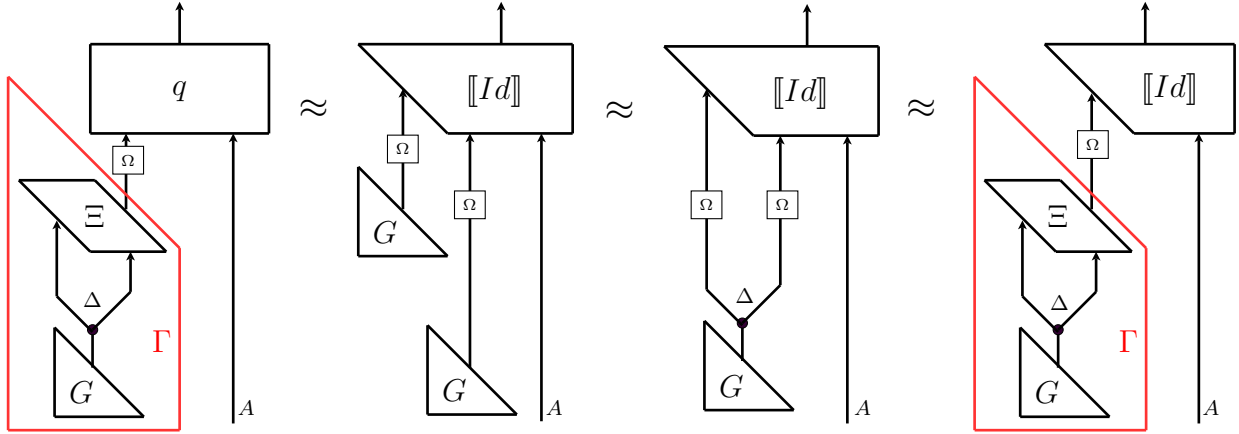
\begin{figure}[ht]
\begin{center}
\newcommand{\universal}{$\Semantics{Id}$}
\renewcommand{\gee}{q}
\renewcommand{\program}{$G$}
\newcommand{\progH}{\color{red}$\Gamma$}
\newcommand{\Branch}{$\scriptstyle \Delta$}
\newcommand{\parteval}{\Xi}
\newcommand{\Beliefs}{$\scriptscriptstyle\DP$}
\newcommand{\Causes}{$\scriptstyle A$}
\newcommand{\EQLS}{\large $\approx$}
\def\JPicScale{.475}
\begin{center}
\ifx\JPicScale\undefined\def\JPicScale{1}\fi
\psset{unit=\JPicScale mm}
\psset{linewidth=0.3,dotsep=1,hatchwidth=0.3,hatchsep=1.5,shadowsize=1,dimen=middle}
\psset{dotsize=0.7 2.5,dotscale=1 1,fillcolor=black}
\psset{arrowsize=1 2,arrowlength=1,arrowinset=0.25,tbarsize=0.7 5,bracketlength=0.15,rbracketlength=0.15}
\begin{pspicture}(0,0)(343.38,120)
\psline[linewidth=0.75](98.12,108.12)(158.12,108.12)
\psline[linewidth=0.75](98.12,73.75)(98.12,48.75)
\psline[linewidth=0.75](158.12,83.12)(158.12,108.12)
\psline[linewidth=0.75]{<-}(138.12,120)(138.12,108.12)
\psline[linewidth=0.75](123.12,83.12)(158.12,83.12)
\psline[linewidth=0.75]{<-}(149.38,83.12)(149.38,2.5)
\psline[linewidth=0.75](98.12,48.75)(123.12,48.75)
\rput(136.24,95){\universal}
\rput(105.62,55.62){\program}
\rput(85.62,90){\EQLS}
\psline[linewidth=0.75](98.12,73.75)(123.12,48.75)
\psline[linewidth=0.75]{<-}(110.62,95.62)(110.62,61.88)
\psline[linewidth=0.75](98.12,108.12)(123.12,83.12)
\psline[linewidth=0.75](23.12,108.12)(73.12,108.12)
\psline[linewidth=0.75](73.12,83.12)(73.12,108.12)
\psline[linewidth=0.75]{<-}(48.12,120)(48.12,108.12)
\psline[linewidth=0.75](23.12,83.12)(73.12,83.12)
\psline[linewidth=0.75]{<-}(63.12,83.12)(63.12,1.88)
\psline[linewidth=0.75]{<-}(33.12,83.12)(33.12,61.87)
\psline[linewidth=0.75](23.12,108.12)(23.12,83.12)
\rput(48.12,95){$\gee$}
\psline[linewidth=0.75](3.12,70)(24.38,70)
\psline[linewidth=0.75](44.38,50)(23.12,50)
\psline[linewidth=0.75](24.38,70)(44.38,50)
\psline[linewidth=0.75](3.12,70)(23.12,50)
\psline[linewidth=0.75](23.75,28.12)(23.75,17.5)
\newrgbcolor{userFillColour}{0.2 0 0.2}
\rput{0}(23.76,28.13){\psellipse[fillcolor=userFillColour,fillstyle=solid](0,0)(1.42,-1.41)}
\psline[linewidth=0.75]{<-}(34.38,50)(34.38,38.75)
\psline[linewidth=0.75]{<-}(13.74,58.75)(13.74,37.5)
\psline[linewidth=0.75](34.38,38.75)(23.75,28.12)
\psline[linewidth=0.75](13.74,37.5)(23.75,27.5)
\psline[linewidth=0.75]{<-}(129.38,83.12)(129.38,17.5)
\psline[linewidth=0.75](13.12,28.75)(13.12,3.75)
\psline[linewidth=0.75](13.12,3.75)(38.12,3.75)
\rput(20.62,10.62){\program}
\psline[linewidth=0.75](13.12,28.75)(38.12,3.75)
\psline[linewidth=0.75](117.5,29.38)(117.5,4.38)
\psline[linewidth=0.75](117.5,4.38)(142.5,4.38)
\rput(125,11.25){\program}
\psline[linewidth=0.75](117.5,29.38)(142.5,4.38)
\psline[linewidth=0.75](183.74,107.5)(243.74,107.5)
\psline[linewidth=0.75](243.74,82.5)(243.74,107.5)
\psline[linewidth=0.75]{<-}(223.74,119.38)(223.74,107.5)
\psline[linewidth=0.75](208.74,82.5)(243.74,82.5)
\psline[linewidth=0.75]{<-}(235,82.5)(235,1.88)
\rput(221.86,94.38){\universal}
\rput(171.24,89.38){\EQLS}
\psline[linewidth=0.75](183.74,107.5)(208.74,82.5)
\psline[linewidth=0.75](205.62,30)(205.62,19.38)
\newrgbcolor{userFillColour}{0.2 0 0.2}
\rput{0}(205.62,30){\psellipse[fillcolor=userFillColour,fillstyle=solid](0,0)(1.42,-1.41)}
\psline[linewidth=0.75]{<-}(216.24,82.5)(216.24,40.62)
\psline[linewidth=0.75]{<-}(195.62,95.62)(195.62,39.38)
\psline[linewidth=0.75](216.24,40.62)(205.62,30)
\psline[linewidth=0.75](195.62,39.38)(205.62,29.38)
\psline[linewidth=0.75](195,30.62)(195,5.62)
\psline[linewidth=0.75](195,5.62)(220,5.62)
\rput(202.5,12.5){\program}
\psline[linewidth=0.75](195,30.62)(220,5.62)
\psline[linewidth=0.75](290.24,108.12)(343.38,108.12)
\psline[linewidth=0.75](343.38,83.12)(343.38,108.12)
\psline[linewidth=0.75]{<-}(318.38,120)(318.38,108.12)
\psline[linewidth=0.75](315.24,83.12)(343.38,83.12)
\psline[linewidth=0.75]{<-}(333.38,83.12)(333.38,1.88)
\psline[linewidth=0.75]{<-}(303.38,94.38)(303.38,61.87)
\psline[linewidth=0.75](290.24,108.12)(315.24,83.12)
\psline[linewidth=0.75](273.38,70)(294.62,70)
\psline[linewidth=0.75](314.62,50)(293.38,50)
\psline[linewidth=0.75](294.62,70)(314.62,50)
\psline[linewidth=0.75](273.38,70)(293.38,50)
\psline[linewidth=0.75](294,28.12)(294,17.5)
\newrgbcolor{userFillColour}{0.2 0 0.2}
\rput{90}(294,28.13){\psellipse[fillcolor=userFillColour,fillstyle=solid](0,0)(1.41,-1.4)}
\psline[linewidth=0.75]{<-}(304.62,50)(304.62,38.75)
\psline[linewidth=0.75]{<-}(284,58.75)(284,37.5)
\psline[linewidth=0.75](304.62,38.75)(294,28.12)
\psline[linewidth=0.75](284,37.5)(294,27.5)
\psline[linewidth=0.75](283.38,28.75)(283.38,3.75)
\psline[linewidth=0.75](283.38,3.75)(308.38,3.75)
\rput(290.88,10.62){\program}
\psline[linewidth=0.75](283.38,28.75)(308.38,3.75)
\rput(325.24,95.62){\universal}
\rput(255.62,90){\EQLS}
\rput(24.38,60){$\parteval$}
\rput(294,60){$\parteval$}
\newrgbcolor{userLineColour}{1 0.2 0.2}
\psline[linewidth=0.75,linecolor=userLineColour](0.12,99)(0.12,0)
\newrgbcolor{userLineColour}{1 0.2 0.2}
\psline[linewidth=0.75,linecolor=userLineColour](0.12,99)(48.12,51)
\rput[r](45.88,25){\progH}
\newrgbcolor{userLineColour}{1 0.2 0.2}
\psline[linewidth=0.75,linecolor=userLineColour](48.12,51)(48.12,0)
\newrgbcolor{userLineColour}{1 0.2 0.2}
\psline[linewidth=0.75,linecolor=userLineColour](0.12,0)(48.12,0)
\newrgbcolor{userLineColour}{1 0.2 0.2}
\psline[linewidth=0.75,linecolor=userLineColour](270.62,99)(270.62,0)
\newrgbcolor{userLineColour}{1 0.2 0.2}
\psline[linewidth=0.75,linecolor=userLineColour](270.62,99)(318.62,51)
\rput[r](316.38,25){\progH}
\newrgbcolor{userLineColour}{1 0.2 0.2}
\psline[linewidth=0.75,linecolor=userLineColour](318.62,51)(318.62,0)
\newrgbcolor{userLineColour}{1 0.2 0.2}
\psline[linewidth=0.75,linecolor=userLineColour](270.62,0)(318.62,0)
\rput(0,0){\rput(303.75,80){\psframebox[fillcolor=white,fillstyle=solid]{\Beliefs}}}
\rput[l](64.38,6.25){\Causes}
\rput[l](150.62,6.25){\Causes}
\rput[l](236.88,6.25){\Causes}
\rput[l](334.38,6.25){\Causes}
\rput(0,0){\rput(110.62,78.75){\psframebox[fillcolor=white,fillstyle=solid]{\Beliefs}}}
\rput(0,0){\rput(129.38,63.12){\psframebox[fillcolor=white,fillstyle=solid]{\Beliefs}}}
\rput(0,0){\rput(195.62,63.12){\psframebox[fillcolor=white,fillstyle=solid]{\Beliefs}}}
\rput(0,0){\rput(216.88,63.12){\psframebox[fillcolor=white,fillstyle=solid]{\Beliefs}}}
\rput[b](206.25,35.62){\Branch}
\rput[b](294.38,33.75){\Branch}
\rput[b](23.75,34.38){\Branch}
\rput(0,0){\rput(33.75,75.62){\psframebox[fillcolor=white,fillstyle=solid]{\Beliefs}}}
\end{pspicture}
\end{center}
\caption{Setting $\Gamma = \Xi\circ \Delta \circ G$ gives $q\circ(\Gamma\ttimes A)\approx \Semantics{\Gamma}$ }
\label{Fig:proof}
\end{center}
\end{figure}

The crucial insight is that the partial model $\Xi:\DP\ttimes \DP\to \DP$ from Sec.~\ref{Sec:partial} induces the $\DP$-parametrized model $\DP\tto{\Delta} \DP\ttimes\DP \tto\Xi \DP$, or in the indexed form $\Xi(\omega, \omega)$, which for every $\omega \in \DP$ predicts the effect of steering the $\DP$-parametrized process $\Semantics \omega$ to $\omega$. This "self-model" $\Xi\circ\Delta$ is displayed in Fig.~\ref{Fig:fixp-1} on the left, where we precompose the given process $q$ with it.
Axiom II now gives a model $G$ of the resulting composite process. Fig.~\ref{Fig:proof} shows that 
$\Gamma = \Xi\circ \Delta \circ G$ provides the claimed causal model of the given process $\DP\ttimes A\tto q B$, parametrized over the possible models. The first step follows from the definition of $G$ in Fig.~\ref{Fig:fixp-1}, by substituting $G$ also as the input. The second step folds two copies of $G$ into one. The third step follows from the definition of $\Xi$ in Fig.~\ref{Fig:specializer}. The upshot is that the causal process $\Semantics{Id}\circ(\Gamma\ttimes A)$ on the right, which is by Sec.~\ref{Sec:universal} just the process $\Semantics\Gamma$ modeled by $\Gamma$, is indistinguishable from the instance $q\circ(\Gamma\ttimes A)$ on the left, obtained by steering the process $q$, which is parametrized  over its models, to the model $\Gamma$.  The causal model $\Gamma$ thus steers the given causal process $q$ to an instance which confirms $\Gamma$ as its model.

%
%
%
%
%
%
%
%
%


\section{Summary: Towards artificial causal cognition? 
}\label{Sec:Outro}

The concept of causality made a remarkable historic voyage, from ancient physics and metaphysics \cite{Aristotle:Physics} to modern physics \cite{BohmD:causality, Ciribella,CoeckeB:CQM-caus}, through  incisive critiques in philosophy \cite{Hume:enquiry,RussellB:cause}, through enduring attention in psychology \cite{Michotte,GopnikA:learning,SlomanS:Book},   all the way to its current promise in machine learning and in AI \cite{koller2009probabilistic,pearl2009causality,pearl2014probabilistic,pearl2018book,PetersJ:inference,spirtes2000causation}. How much does the new outlook change the scene of causal cognition? 

If subjectivity is implied by cognitive self-confirmation, then the results presented in this paper suggest that all causal cognition, human or artificial, is inexorably subjective, as soon as models cannot be eliminated as causal factors. In particular, we have shown how causal models and causal processes can be steered to support each other, and adapted to absorb any new evidence. 
This is a concerning conclusion, if scientific theories are required to be falsifiable. 
The fact that causal cognition, as modeled here, permits steering and self-validation under the assumptions as mild as axioms I--III, can also be interpreted as demonstrating that causal cognition cannot be understood just in terms of logic and structure.  Such a conclusion would have significant repercussions on the AI research.

Computer science has been recognized as a natural science almost from the outset.
 This is by now largely accepted in principle, although often misunderstood in the popular narrative, and dismissed as irrelevant by many working computer scientists and engineers. 
With the advent of the web, and the spread of personal devices, computation propagated through the vast area of economic, political, and social processes, and came to play a role in many forms of human cognition. Changing the angle only slightly, the same process can be viewed as evolution of artificial cognition within a natural process of network computation.
%
%
%
Whether the androids of the future will be attracted to causal illusions of action movies remains to be seen; but the humanities of the present cannot avoid studying human cognition together with artificial cognition. Like identical twins, or like parallel universes, or like Lao Tse and the butterfly, the two can be reliably distinguished only by themselves, observing each other from within, but difficult to tell apart for the external observers. The human gave rise to the artificial, and then the artificial gave rise to a new human, and then maybe there was a new artificial, and then they lost count.


\paragraph{Psychology of artificial mind?} Herbert Simon, one of the originators AI, anticipated the creation of a whole new realm of \emph{"The Sciences of the Artificial"} \cite{SimonH:artificial}. With the AI technologies permeating market and politics, a \emph{psychology of the artificial}\/ seems to have become not just a real possibility, but also an urgent task.

%
%
%
%
%
%
%
%
%


\addcontentsline{toc}{part}{References}
\bibliography{cogsci-ref,philosophy,PavlovicD,CT}
\bibliographystyle{plain}

\appendix
\addcontentsline{toc}{part}{\appendixname}
\label{Sec:Appendix}

\clearpage

{\Large \bf Appendix}

\section{Composing and decomposing {\causations} }
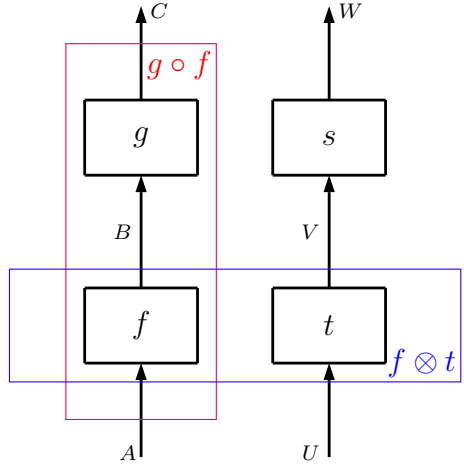
\begin{figure}[ht]
\begin{center}
\newcommand{\machine}{$f$}
\renewcommand{\gee}{$g$}
\newcommand{\kee}{$s$}
\renewcommand{\hee}{t}
\newcommand{\nameslang}{\scriptstyle B}
\newcommand{\seqcompp}{{\color{red}g\circ f}}
\newcommand{\parcompp}{{\color{blue}f\ttimes t}}
\newcommand{\inputt}{\scriptstyle A} 
\newcommand{\outpt}{$\scriptstyle C$}
\newcommand{\otherinputt}{\scriptstyle U}
\newcommand{\otheroutpt}{\scriptstyle V} 
\newcommand{\outpttt}{$\scriptstyle W$}
\def\JPicScale{.5}
\ifx\JPicScale\undefined\def\JPicScale{1}\fi
\psset{unit=\JPicScale mm}
\psset{linewidth=0.3,dotsep=1,hatchwidth=0.3,hatchsep=1.5,shadowsize=1,dimen=middle}
\psset{dotsize=0.7 2.5,dotscale=1 1,fillcolor=black}
\psset{arrowsize=1 2,arrowlength=1,arrowinset=0.25,tbarsize=0.7 5,bracketlength=0.15,rbracketlength=0.15}
\begin{pspicture}(0,0)(115,115)
\psline[linewidth=0.75](15,40)(45,40)
\psline[linewidth=0.75](15,40)(15,20)
\psline[linewidth=0.75](45,20)(45,40)
\psline[linewidth=0.75,arrowsize=1.5 2,arrowlength=1.5,arrowinset=0]{<-}(30,70)(30,40)
\rput[r](27.5,55){$\nameslang$}
\rput[r](28.75,-3.75){$\inputt$}
\psline[linewidth=0.75](15,20)(45,20)
\psline[linewidth=0.75,arrowsize=1.5 2,arrowlength=1.5,arrowinset=0]{<-}(30,20)(30,-5)
\rput(30,30){\machine}
\psline[linewidth=0.75](15,90)(45,90)
\psline[linewidth=0.75](15,90)(15,70)
\psline[linewidth=0.75](45,70)(45,90)
\psline[linewidth=0.75,arrowsize=1.5 2,arrowlength=1.5,arrowinset=0]{<-}(30,115)(30,90)
\psline[linewidth=0.75](15,70)(45,70)
\rput(30,80){\gee}
\rput[l](32.5,113.75){\outpt}
\psline[linewidth=0.75](65,40)(95,40)
\psline[linewidth=0.75](65,40)(65,20)
\psline[linewidth=0.75](95,20)(95,40)
\psline[linewidth=0.75,arrowsize=1.5 2,arrowlength=1.5,arrowinset=0]{<-}(80,70)(80,40)
\psline[linewidth=0.75](65,20)(95,20)
\psline[linewidth=0.75,arrowsize=1.5 2,arrowlength=1.5,arrowinset=0]{<-}(80,20)(80,-5)
\rput[r](77.5,-3.75){$\otherinputt$}
\rput[r](77.5,55){$\otheroutpt$}
\rput(80,30){$\hee$}
\newrgbcolor{userLineColour}{1 0 0.4}
\pspolygon[linecolor=userLineColour](10,105)(50,105)(50,5)(10,5)
\newrgbcolor{userLineColour}{0.2 0 1}
\pspolygon[linecolor=userLineColour](-5,45)(115,45)(115,15)(-5,15)
\rput[br](113.75,16.25){$\parcompp$}
\rput[tr](48.75,103.12){$\seqcompp$}
\psline[linewidth=0.75](65,90)(65,70)
\psline[linewidth=0.75](95,70)(95,90)
\psline[linewidth=0.75,arrowsize=1.5 2,arrowlength=1.5,arrowinset=0]{<-}(80,115)(80,90)
\psline[linewidth=0.75](65,70)(95,70)
\psline[linewidth=0.75](65,90)(95,90)
\rput(80,80){\kee}
\rput[l](82.5,113.75){\outpttt}
\end{pspicture}
\vspace{.5\baselineskip}
\caption{Sequential composition $g\circ f$ and parallel composition $f\ttimes t$}
\label{Fig:godement}
\end{center}
\end{figure}

The salient feature of the string diagram presentation of processes is that the two dimensions of string diagrams correspond to two kinds of function composition. This is displayed in Fig.~\ref{Fig:godement}.
\begin{itemize}
\item \emph{\textbf{Sequential composition}} of {\causations} corresponds to linking the corresponding string diagrams \emph{vertically}: 
the effects of the {\causation} $A\tto f B$ are passed to the {\causation} $B\tto g C$ to produce the {\causation} $A\tto{g\circ f} C$;
\item \emph{\textbf{parallel composition}}\/ lays {\causations} next to each other \emph{horizontally}: 
the processes $A\tto f B$ and $U\tto t V$ are kept independent, and their composite $A\ttimes U\tto{f\ttimes t} B\ttimes V$ takes each processes causal factors separately, and produces their effects without any interactions between the two:
\end{itemize}
As categorical structures, these operations are captured as the mappings
\bear
\CCC(A,B) \times \CCC(B,C) & \tto\circ & \CCC(A,C)\\
\CCC(A,B) \times \CCC(U,V) & \tto\ttimes & \CCC(A\ttimes U,B\ttimes V)
\eear
\paragraph{Meaning of the sequential composition.} The composite $A\tto{g\circ f}C$ inputs the cause $a\in A$ and outputs the effect $g(f(a))\in C$ of the cause $f(a)\in B$, which is itself the effect of the cause $a\in A$ . In summary, we have
\begin{gather}\label{eq:circ}
\prooftree
\prooftree
a\in A\qquad f:A\to B
\justifies
f(a) \in B 
\endprooftree 
\quad
g: B\to C
\justifies
g\circ f(a)\ =\ g(f(a))\in C
\endprooftree
\end{gather}
 
 \paragraph{Meaning of the parallel composition and product types.} Since the strings in a string diagram correspond to types, drawing parallel strings leads to product types, like $A\ttimes U$, which is the name of the type corresponding to the strings $A$ and $U$ running in parallel. The  events of this type are the pairs $<a,u>\in A\ttimes U$, where $a\in A$ and $u\in U$. The parallel composite $A\ttimes U\tto{f\ttimes t} B\ttimes V$ can thus be defined as the {\causation} transforming independent pairs of causes into independent pairs of effects, without any interferences between the components:
\[
\prooftree
\prooftree
\prooftree
<a,u>\in A\ttimes U
\justifies
a\in A
\endprooftree \qquad f:A\to B
\justifies
f(a) \in B
\endprooftree 
\qquad
\prooftree
\prooftree
<a,u>\in A\ttimes U
\justifies
u\in U
\endprooftree \qquad  t: U\to V
\justifies
t(u) \in V
\endprooftree
\justifies
(f\ttimes t)<a,u> = \big<f(a), t(u)\big> \in B\ttimes V
\endprooftree
\]

\section{Units}
\begin{figure}[htbp]
\begin{center}
\newcommand{\inputt}{\scriptstyle A} 
\newcommand{\ahh}{\scriptstyle c}
\newcommand{\eex}{\scriptstyle s}
\newcommand{\bee}{\scriptstyle e} 
\def\JPicScale{.5}
\ifx\JPicScale\undefined\def\JPicScale{1}\fi
\psset{unit=\JPicScale mm}
\psset{linewidth=0.3,dotsep=1,hatchwidth=0.3,hatchsep=1.5,shadowsize=1,dimen=middle}
\psset{dotsize=0.7 2.5,dotscale=1 1,fillcolor=black}
\psset{arrowsize=1 2,arrowlength=1,arrowinset=0.25,tbarsize=0.7 5,bracketlength=0.15,rbracketlength=0.15}
\begin{pspicture}(0,0)(120,30)
\psline[linewidth=0.75](110,30)(100,20)
\psline[linewidth=0.75](120,20)(110,30)
\rput[r](107.5,2.5){$\inputt$}
\psline[linewidth=0.75](100,20)(120,20)
\psline[linewidth=0.75,arrowsize=1.5 2,arrowlength=1.5,arrowinset=0]{<-}(110,20)(110,0)
\psline[linewidth=0.75](20,10)(10,0)
\psline[linewidth=0.75](10,0)(0,10)
\rput[l](12.5,27.5){$\inputt$}
\psline[linewidth=0.75](0,10)(20,10)
\psline[linewidth=0.75,arrowsize=1.5 2,arrowlength=1.5,arrowinset=0]{<-}(10,30)(10,10)
\psline[linewidth=0.75](70,15)(60,5)
\psline[linewidth=0.75](60,5)(50,15)
\psline[linewidth=0.75](70,15)(60,25)
\psline[linewidth=0.75](60,25)(50,15)
\rput(60,15){$\eex$}
\rput(110,24.38){$\ahh$}
\rput(10.62,5.62){$\bee$}
\end{pspicture}
\caption{String diagrams with invisible strings}
\label{Fig:vector-scalar}
\end{center}
\end{figure}
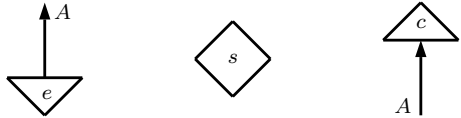

\paragraph{Vectors, scalars, covectors.} There are processes where events occur with no visible causes; and  there are processes where events have no visible effects. Such processes correspond, respectively, to string diagrams $c$ and $e$ in Fig.~\ref{Fig:vector-scalar}. There are even processes with no observable causes or effects, like the one represented by the diamond $s$ in the middle of Fig.~\ref{Fig:vector-scalar}. When there are no strings at the bottom, or at the top of a box in a string diagram, we usually contract the bottom side, or the top side, into a point, and the box becomes a triangle. When there are no strings either at the bottom or at the top, then contracting both of them results in a diamond. A diamond with no inputs and outputs may still contain a lot of information. E.g., if {\causations} are timed, then reading the time without observing anything else would be a process with no observable causes or effects. 
If processes are viewed as linear operators, then  those that do not consume any input vectors and do not produce any output vectors are scalars. The processes that do not consume anything but produce output are just vectors, since they boil down to their output. The processes that do not produce vectors, but only consume them, are just covectors, or linear functionals.

\textbf{Invisible string: The unit type $I$.} Since every process must have an input type and an output type for formal reasons, in order to fit into a category of processes, a process that does not have any actual causes is presented in the form $I\tto e A$, where $I$ is the \emph{unit type}, satisfying
\beq\label{eq:I} I\ttimes A \ =\ A \ =\ A\ttimes I\eeq
for every type $A$. The unit type is thus degenerate, in the sense that any number of its copies can be added to any type, without changing its elements. It is easy to see that it is unique, as the units in algebra tend to be. A process that does not output any effects is then in the form $A\tto c I$. The unit type is the unit with respect to the type product and to the parallel composition of processes, just like 0 is the unit with respect to the addition of numbers. It can be thought of as the type of a single event that never interferes with any other events. 

Since it is introduced only to make sure that everything has a type, but otherwise has no visible causes or effects, the unit type is usually not drawn in string diagrams, but thought of as an \emph{"invisible string"}. In Fig.~\ref{Fig:vector-scalar}, the invisible string is coming in below $e$, and out above $c$, and on both sides of $s$.

%
%
%
%
%

%

\paragraph{Invisible boxes: The unit processes $\id_A$.} For every type $A$ there is a unit process $A\tto{\id_A} A$, called the \emph{identity}\/ of $A$, such that
\beq\label{eq:id} \id_B\circ f\ =\ f\ =\ f\circ \id_A\eeq
holds for every process $A\tto f B$. This property is clearly analogous to \eqref{eq:I}, but has a different meaning: the {\causation} $\id_A$ inputs causes and outputs effects, both of type $A$; but it does not modify anything: it just outputs every cause $a\in A$ as its own effect. Since the box corresponding to $\id_A$ thus just passes to the output string whatever comes at the input string, and there is nothing else in it, we do not draw such a box. In string diagrams, the boxes corresponding to the identity processes $\id_A$ can thus be thought of as \emph{"invisible boxes"}. In a sense, the strings themselves play the role of the identities.  

Because of \eqref{eq:I}, any number of invisible strings can be added to any string diagram without changing its meaning. Because of \eqref{eq:id}, any number of invisible boxes can be added on any string, visible or invisible, without changing its meaning. This justifies eliding the units not only from string diagrams, but also from algebraic expressions, and writing things like
\[
f\ttimes U\  =\  f\ttimes \id_U \qquad \mbox{ and } \qquad A\ttimes t\  =\  \id_A \ttimes f 
\]
With this notation, the algebra of tensor products, studied in serious math, boils down to a single law, from which everything else follows:

\section{The middle-two-interchange law} The main reason why string diagrams are convenient is that their geometry captures the \emph{middle-two-interchange}\/ law:
\bea\label{eq:godementLine}
(f;g)\ttimes(t;s) & = & (f\ttimes t);(g\ttimes s)
\eea
Note that the both sides of this equation correspond to the same string diagram, displayed in Fig.~\ref{Fig:godement}. The left-hand side of \eqref{eq:godementLine} is obtained by first reading the sequential compositions vertically, and then composing them in parallel horizontally; whereas the right-hand side is obtained by first reading the parallel compositions horizontally, and then composing them in sequence vertically. Sliding boxes along stings provides an easy way to derive further equations from the middle-two-interchange low of \eqref{eq:godementLine}, such as
\[
(f\ttimes t);(g\ttimes V) \qquad = \qquad (f\ttimes U);(B\ttimes t);(g\ttimes V) \qquad = \qquad (f\ttimes U);(g\ttimes t)
\]
In string diagrams, this is just
\[
\newcommand{\machine}{$f$}
\renewcommand{\gee}{$g$}
\renewcommand{\hee}{t}
\newcommand{\nameslang}{\scriptstyle B}
\newcommand{\seqcompp}{\scriptstyle (f;g)}
\newcommand{\parcompp}{\scriptstyle f\ttimes h}
\newcommand{\inputt}{\scriptstyle A} 
\newcommand{\outpt}{$\scriptstyle C$}
\newcommand{\otherinputt}{\scriptstyle U}
\newcommand{\otheroutpt}{\scriptstyle V} 
\def\JPicScale{.35}
\ifx\JPicScale\undefined\def\JPicScale{1}\fi
\psset{unit=\JPicScale mm}
\psset{linewidth=0.3,dotsep=1,hatchwidth=0.3,hatchsep=1.5,shadowsize=1,dimen=middle}
\psset{dotsize=0.7 2.5,dotscale=1 1,fillcolor=black}
\psset{arrowsize=1 2,arrowlength=1,arrowinset=0.25,tbarsize=0.7 5,bracketlength=0.15,rbracketlength=0.15}
\begin{pspicture}(0,0)(80,55)
\psline[linewidth=0.75](0,-15)(30,-15)
\psline[linewidth=0.75](0,-15)(0,-35)
\psline[linewidth=0.75](30,-35)(30,-15)
\psline[linewidth=0.75,arrowsize=1.5 2,arrowlength=1.5,arrowinset=0]{<-}(15,15)(15,-15)
\rput[r](12.5,0){$\nameslang$}
\rput[r](12.5,-52.5){$\inputt$}
\psline[linewidth=0.75](0,-35)(30,-35)
\psline[linewidth=0.75,arrowsize=1.5 2,arrowlength=1.5,arrowinset=0]{<-}(15,-35)(15,-55)
\rput(15,-25){\machine}
\psline[linewidth=0.75](0,35)(30,35)
\psline[linewidth=0.75](0,35)(0,15)
\psline[linewidth=0.75](30,15)(30,35)
\psline[linewidth=0.75,arrowsize=1.5 2,arrowlength=1.5,arrowinset=0]{<-}(15,55)(15,35)
\psline[linewidth=0.75](0,15)(30,15)
\rput(15,25){\gee}
\rput[l](17.5,52.5){\outpt}
\psline[linewidth=0.75](50,-15)(80,-15)
\psline[linewidth=0.75](50,-15)(50,-35)
\psline[linewidth=0.75](80,-35)(80,-15)
\psline[linewidth=0.75,arrowsize=1.5 2,arrowlength=1.5,arrowinset=0]{<-}(65,55)(65,-15)
\psline[linewidth=0.75](50,-35)(80,-35)
\psline[linewidth=0.75,arrowsize=1.5 2,arrowlength=1.5,arrowinset=0]{<-}(65,-35)(65,-55)
\rput[r](62.5,-52.5){$\otherinputt$}
\rput[l](67.5,52.5){$\otheroutpt$}
\rput(65,-25){$\hee$}
\end{pspicture}

\qquad\quad=\quad\qquad
\ifx\JPicScale\undefined\def\JPicScale{1}\fi
\psset{unit=\JPicScale mm}
\psset{linewidth=0.3,dotsep=1,hatchwidth=0.3,hatchsep=1.5,shadowsize=1,dimen=middle}
\psset{dotsize=0.7 2.5,dotscale=1 1,fillcolor=black}
\psset{arrowsize=1 2,arrowlength=1,arrowinset=0.25,tbarsize=0.7 5,bracketlength=0.15,rbracketlength=0.15}
\begin{pspicture}(0,0)(80,55)
\psline[linewidth=0.75](0,-15)(30,-15)
\psline[linewidth=0.75](0,-15)(0,-35)
\psline[linewidth=0.75](30,-35)(30,-15)
\psline[linewidth=0.75,arrowsize=1.5 2,arrowlength=1.5,arrowinset=0]{<-}(15,15)(15,-15)
\rput[r](12.5,0){$\nameslang$}
\rput[r](12.5,-52.5){$\inputt$}
\psline[linewidth=0.75](0,-35)(30,-35)
\psline[linewidth=0.75,arrowsize=1.5 2,arrowlength=1.5,arrowinset=0]{<-}(15,-35)(15,-55)
\rput(15,-25){\machine}
\psline[linewidth=0.75](0,35)(30,35)
\psline[linewidth=0.75](0,35)(0,15)
\psline[linewidth=0.75](30,15)(30,35)
\psline[linewidth=0.75,arrowsize=1.5 2,arrowlength=1.5,arrowinset=0]{<-}(15,55)(15,35)
\psline[linewidth=0.75](0,15)(30,15)
\rput(15,25){\gee}
\rput[l](17.5,52.5){\outpt}
\psline[linewidth=0.75](50,10)(80,10)
\psline[linewidth=0.75](50,10)(50,-10)
\psline[linewidth=0.75](80,-10)(80,10)
\psline[linewidth=0.75,arrowsize=1.5 2,arrowlength=1.5,arrowinset=0]{<-}(65,55)(65,10)
\psline[linewidth=0.75](50,-10)(80,-10)
\psline[linewidth=0.75,arrowsize=1.5 2,arrowlength=1.5,arrowinset=0]{<-}(65,-10)(65,-55)
\rput[r](62.5,-52.5){$\otherinputt$}
\rput[l](67.5,52.5){$\otheroutpt$}
\rput(65,0){$\hee$}
\end{pspicture}

\qquad\quad=\quad\qquad
\ifx\JPicScale\undefined\def\JPicScale{1}\fi
\psset{unit=\JPicScale mm}
\psset{linewidth=0.3,dotsep=1,hatchwidth=0.3,hatchsep=1.5,shadowsize=1,dimen=middle}
\psset{dotsize=0.7 2.5,dotscale=1 1,fillcolor=black}
\psset{arrowsize=1 2,arrowlength=1,arrowinset=0.25,tbarsize=0.7 5,bracketlength=0.15,rbracketlength=0.15}
\begin{pspicture}(0,0)(80,55)
\psline[linewidth=0.75](0,-15)(30,-15)
\psline[linewidth=0.75](0,-15)(0,-35)
\psline[linewidth=0.75](30,-35)(30,-15)
\psline[linewidth=0.75,arrowsize=1.5 2,arrowlength=1.5,arrowinset=0]{<-}(15,15)(15,-15)
\rput[r](12.5,0){$\nameslang$}
\rput[r](12.5,-52.5){$\inputt$}
\psline[linewidth=0.75](0,-35)(30,-35)
\psline[linewidth=0.75,arrowsize=1.5 2,arrowlength=1.5,arrowinset=0]{<-}(15,-35)(15,-55)
\rput(15,-25){\machine}
\psline[linewidth=0.75](0,35)(30,35)
\psline[linewidth=0.75](0,35)(0,15)
\psline[linewidth=0.75](30,15)(30,35)
\psline[linewidth=0.75,arrowsize=1.5 2,arrowlength=1.5,arrowinset=0]{<-}(15,55)(15,35)
\psline[linewidth=0.75](0,15)(30,15)
\rput(15,25){\gee}
\rput[l](17.5,52.5){\outpt}
\psline[linewidth=0.75](50,35)(80,35)
\psline[linewidth=0.75](50,35)(50,15)
\psline[linewidth=0.75](80,15)(80,35)
\psline[linewidth=0.75,arrowsize=1.5 2,arrowlength=1.5,arrowinset=0]{<-}(65,55)(65,35)
\psline[linewidth=0.75](50,15)(80,15)
\psline[linewidth=0.75,arrowsize=1.5 2,arrowlength=1.5,arrowinset=0]{<-}(65,15)(65,-55)
\rput[r](62.5,-52.5){$\otherinputt$}
\rput[l](67.5,52.5){$\otheroutpt$}
\rput(65,25){$\hee$}
\end{pspicture}

\]
\vspace{3\baselineskip}

\section{Functions}\label{Sec:functions}
Causations are usually uncertain to some extent: a cause induces an effect with a certain probability. Those causations that are certain, so that the input always produces the output, and the same input always produces the same output, are called \emph{functions}. If causations are presented as stochastic processes, then functions are the subfamily deterministic processes. 

An intrinsic characterization of functions in a monoidal framewrok is provided by the following definitions.

A \textbf{data type} is a  type $A\in \CCC$, equipped with \emph{data services}, i.e. the operations
\beq\label{eq:comon}
I \oot\cun A \tto \cmn A\ttimes A
\eeq
respectively called \emph{deleting} and \emph{copying}, which satisfy the following equations:

\begin{alignat*}{7}
\comp {\cmn}{(\cmn \ttimes A)}  &\ \ =\ \   \comp{\cmn}{(A\ttimes \cmn)} &&\qquad&\qquad&\quad&
\comp{\cmn}{(\cun \ttimes A)}  &\ \  =\ \  &\   \id_A  &\ \   =\ \  &\ \  \comp{\cmn}{(A\ttimes \cun)}
\\[1ex]
\def\JPicScale{.85} 
\ifx\JPicScale\undefined\def\JPicScale{1}\fi
\psset{unit=\JPicScale mm}
\psset{linewidth=0.3,dotsep=1,hatchwidth=0.3,hatchsep=1.5,shadowsize=1,dimen=middle}
\psset{dotsize=0.7 2.5,dotscale=1 1,fillcolor=black}
\psset{arrowsize=1 2,arrowlength=1,arrowinset=0.25,tbarsize=0.7 5,bracketlength=0.15,rbracketlength=0.15}
\begin{pspicture}(0,0)(15,7.5)
\psline(5.62,-0.62)
(10.62,-5.62)
(9.38,-5.62)(10,-5.62)
\rput{0}(4.38,0.62){\psellipse[fillstyle=solid](0,0)(1.56,-1.57)}
\rput{0}(10.32,-5.31){\psellipse[fillstyle=solid](0,0)(1.56,-1.57)}
\psline(15,7.5)
(15,5)
(15,-0.63)(11.25,-4.38)
\psline(9.37,7.5)
(9.37,6.87)
(9.38,5)(5.63,1.25)
\psline(-0.63,7.5)
(-0.63,6.87)
(-0.62,5)(3.13,1.25)
\psline(10,-6.88)
(10,-10)
(10,-10.62)(10,-10)
\end{pspicture}
\   &\ \ =\ \   \def\JPicScale{.85} 
\ifx\JPicScale\undefined\def\JPicScale{1}\fi
\psset{unit=\JPicScale mm}
\psset{linewidth=0.3,dotsep=1,hatchwidth=0.3,hatchsep=1.5,shadowsize=1,dimen=middle}
\psset{dotsize=0.7 2.5,dotscale=1 1,fillcolor=black}
\psset{arrowsize=1 2,arrowlength=1,arrowinset=0.25,tbarsize=0.7 5,bracketlength=0.15,rbracketlength=0.15}
\begin{pspicture}(0,0)(15.63,8.12)
\rput{0}(10.62,0.62){\psellipse[fillstyle=solid](0,0)(1.56,-1.56)}
\psline(15.62,8.12)
(15.62,7.5)
(15.63,5.62)(11.88,1.87)
\psline(5.62,8.12)
(5.62,7.5)
(5.63,5.62)(9.38,1.87)
\rput{0}(4.38,-5.62){\psellipse[fillstyle=solid](0,0)(1.57,-1.57)}
\psline(-0.63,7.5)
(-0.62,1.25)
(-0.62,-0.63)(3.13,-4.38)
\pscustom[]{\psline(9.38,-0.62)(4.38,-5.62)
\psbezier(4.38,-5.62)(4.38,-5.62)(4.38,-5.62)
\psbezier(4.38,-5.62)(4.38,-5.62)(4.38,-5.62)
}
\psline(4.38,-6.88)
(4.38,-10)
(4.38,-10.62)(4.38,-10)
\end{pspicture}
 &&&&& 
\def\JPicScale{.85} 
\ifx\JPicScale\undefined\def\JPicScale{1}\fi
\psset{unit=\JPicScale mm}
\psset{linewidth=0.3,dotsep=1,hatchwidth=0.3,hatchsep=1.5,shadowsize=1,dimen=middle}
\psset{dotsize=0.7 2.5,dotscale=1 1,fillcolor=black}
\psset{arrowsize=1 2,arrowlength=1,arrowinset=0.25,tbarsize=0.7 5,bracketlength=0.15,rbracketlength=0.15}
\begin{pspicture}(0,0)(10,8.12)
\psline(0.62,0)
(5.62,-5)
(4.38,-5)(5,-5)
\rput{0}(-0.62,1.25){\psellipse[fillstyle=solid](0,0)(1.57,-1.56)}
\rput{0}(5.31,-4.69){\psellipse[fillstyle=solid](0,0)(1.57,-1.56)}
\psline(10,8.12)
(10,5.62)
(10,0)(6.25,-3.75)
\psline(5,-6.25)
(5,-10)
(5,-10.62)(5,-10)
\end{pspicture}
&\ \ =\ \ & \def\JPicScale{.85} 
\ifx\JPicScale\undefined\def\JPicScale{1}\fi
\psset{unit=\JPicScale mm}
\psset{linewidth=0.3,dotsep=1,hatchwidth=0.3,hatchsep=1.5,shadowsize=1,dimen=middle}
\psset{dotsize=0.7 2.5,dotscale=1 1,fillcolor=black}
\psset{arrowsize=1 2,arrowlength=1,arrowinset=0.25,tbarsize=0.7 5,bracketlength=0.15,rbracketlength=0.15}
\begin{pspicture}(0,0)(0,8.74)
\psline(0,8.74)
(0,-9.38)
(0,-10)(0,-9.38)
\end{pspicture}
\ \   &\ \ =\ \ &  \def\JPicScale{.85} 
\ifx\JPicScale\undefined\def\JPicScale{1}\fi
\psset{unit=\JPicScale mm}
\psset{linewidth=0.3,dotsep=1,hatchwidth=0.3,hatchsep=1.5,shadowsize=1,dimen=middle}
\psset{dotsize=0.7 2.5,dotscale=1 1,fillcolor=black}
\psset{arrowsize=1 2,arrowlength=1,arrowinset=0.25,tbarsize=0.7 5,bracketlength=0.15,rbracketlength=0.15}
\begin{pspicture}(0,0)(12.19,8.12)
\rput{0}(10.62,1.25){\psellipse[fillstyle=solid](0,0)(1.57,-1.56)}
\rput{0}(4.38,-5){\psellipse[fillstyle=solid](0,0)(1.57,-1.56)}
\psline(4.38,-6.25)
(4.38,-10)
(4.38,-10.62)(4.38,-10)
\psline(-0.63,8.12)
(-0.62,1.87)
(-0.62,0)(3.13,-3.75)
\pscustom[]{\psline(9.38,0)(4.38,-5)
\psbezier(4.38,-5)(4.38,-5)(4.38,-5)
\psbezier(4.38,-5)(4.38,-5)(4.38,-5)
}
\end{pspicture}

\end{alignat*}
%
%
%
\bear
\cmn\ \  & = & \varsigma \circ \cmn\\[1.5ex]
\def\JPicScale{.85} 
\ifx\JPicScale\undefined\def\JPicScale{1}\fi
\psset{unit=\JPicScale mm}
\psset{linewidth=0.3,dotsep=1,hatchwidth=0.3,hatchsep=1.5,shadowsize=1,dimen=middle}
\psset{dotsize=0.7 2.5,dotscale=1 1,fillcolor=black}
\psset{arrowsize=1 2,arrowlength=1,arrowinset=0.25,tbarsize=0.7 5,bracketlength=0.15,rbracketlength=0.15}
\begin{pspicture}(0,0)(5,10)
\rput{0}(2.5,-2.5){\psellipse[fillstyle=solid](0,0)(1.57,-1.56)}
\psline(2.5,-2.5)(2.5,-7.5)
\psline(5,0)(2.5,-2.5)
\psline(0,0)(2.5,-2.5)
\psline(5,10)(5,0)
\psline(0,10)(0,0)
\end{pspicture}
 & = & \def\JPicScale{.85} 
\ifx\JPicScale\undefined\def\JPicScale{1}\fi
\psset{unit=\JPicScale mm}
\psset{linewidth=0.3,dotsep=1,hatchwidth=0.3,hatchsep=1.5,shadowsize=1,dimen=middle}
\psset{dotsize=0.7 2.5,dotscale=1 1,fillcolor=black}
\psset{arrowsize=1 2,arrowlength=1,arrowinset=0.25,tbarsize=0.7 5,bracketlength=0.15,rbracketlength=0.15}
\begin{pspicture}(0,0)(5,10)
\rput{0}(2.5,-2.5){\psellipse[fillstyle=solid](0,0)(1.56,-1.57)}
\psline(2.5,-2.5)(2.5,-7.5)
\psline(5,0)(2.5,-2.5)
\psline(0,0)(2.5,-2.5)
\psline(5,2.5)(5,0)
\psline(0,2.5)(0,0)
\psline(0,7.5)(5,2.5)
\psline(0,2.5)(5,7.5)
\psline(5,10)(5,7.5)
\psline(0,10)(0,7.5)
\end{pspicture}

\eear
\vspace{1\baselineskip}

We define \emph{data} as the elements preserved by data services. Such elements are precisely those that can be manipulated using variables \cite{PavlovicD:MSCS97,PavlovicD:Qabs12}.

\paragraph{Remark.} If we dualize data services, i.e. reverse the arrows in \eqref{eq:comon}, we get a binary operation and a constant. Transferring the equations from deleting and copying makes the binary operation associative and commutative, and it makes the constant into the unit. The dual of  data services is thus the structure of a \emph{commutative monoid}. The structure of data services itself is thus a commutative \emph{co}\/monoid.

\paragraph{Functions} are causations that map data to data. A causation $A\tto f B$ is a function if it is \emph{total}\/ and \emph{single-valued}, which respectively corresponds to the two equations in Fig.~\ref{fig-comonoid}. They also make $f$ into a comonoid homomorphism from the data service comonoid on $A$ to the data service comonoid on $B$.

\begin{figure}[htbp]
\begin{center}
\newcommand{\monoidd}{\cmn}
\newcommand{\fun}{\scriptstyle f}
\newcommand{\One}{A}
\newcommand{\Two}{B}
\newcommand{\delete}{\cun}
\def\JPicScale{1.1}
\ifx\JPicScale\undefined\def\JPicScale{1}\fi
\psset{unit=\JPicScale mm}
\psset{linewidth=0.3,dotsep=1,hatchwidth=0.3,hatchsep=1.5,shadowsize=1,dimen=middle}
\psset{dotsize=0.7 2.5,dotscale=1 1,fillcolor=black}
\psset{arrowsize=1 2,arrowlength=1,arrowinset=0.25,tbarsize=0.7 5,bracketlength=0.15,rbracketlength=0.15}
\begin{pspicture}(0,0)(69.06,19.38)
\rput(11.88,10){$=$}
\pspolygon[](1.88,7.5)(5.62,7.5)(5.62,3.75)(1.88,3.75)
\pspolygon[](16.88,15.01)(20.63,15.01)(20.63,11.25)(16.88,11.25)
\pspolygon[](25.62,15.01)(29.39,15.01)(29.39,11.25)(25.62,11.25)
\rput(3.75,5.62){$\fun$}
\rput(18.75,13.12){$\fun$}
\rput(27.5,13.12){$\fun$}
\rput(61.88,10){$=$}
\pspolygon[](51.88,9.38)(55.62,9.38)(55.62,5.62)(51.88,5.62)
\rput(53.76,7.5){$\fun$}
\rput{0}(3.75,12.5){\psellipse[fillstyle=solid](0,0)(1.56,-1.56)}
\pscustom[]{\psline(3.75,8.12)(3.75,7.5)
\psbezier(3.75,7.5)(3.75,7.5)(3.75,7.5)
\psline(3.75,7.5)(3.75,11.25)
}
\pscustom[]{\psline(3.75,0)(3.75,-0.62)
\psbezier(3.75,-0.62)(3.75,-0.62)(3.75,-0.62)
\psline(3.75,-0.62)(3.75,3.75)
}
\psline(3.75,11.88)
(0.62,15)
(-0.62,16.25)(-0.62,19.38)
\psline(4.38,12.5)
(8.12,16.25)
(8.12,17.5)(8.12,19.38)
\rput{0}(23.12,4.38){\psellipse[fillstyle=solid](0,0)(1.57,-1.57)}
\pscustom[]{\psline(23.12,0)(23.12,-0.62)
\psbezier(23.12,-0.62)(23.12,-0.62)(23.12,-0.62)
\psline(23.12,-0.62)(23.12,3.12)
}
\psline(23.12,3.75)
(20,6.88)
(18.75,8.12)(18.75,11.25)
\psline(23.75,4.38)
(27.5,8.12)
(27.5,9.38)(27.5,11.25)
\pscustom[]{\psline(18.75,15)(18.75,15.62)
\psbezier(18.75,15.62)(18.75,15.62)(18.75,15.62)
\psline(18.75,15.62)(18.75,19.38)
}
\pscustom[]{\psbezier(27.5,15)(27.5,15)(27.5,15)(27.5,15)
\psbezier(27.5,15)(27.5,15)(27.5,15)
\psline(27.5,15)(27.5,19.38)
}
\pscustom[]{\psbezier(53.75,0)(53.75,0)(53.75,0)(53.75,0)
\psbezier(53.75,0)(53.75,0)(53.75,0)
\psline(53.75,0)(53.75,5.62)
}
\rput{90}(53.75,14.38){\psellipse[fillstyle=solid](0,0)(1.57,-1.56)}
\pscustom[]{\psline(53.75,10)(53.75,9.38)
\psbezier(53.75,9.38)(53.75,9.38)(53.75,9.38)
\psline(53.75,9.38)(53.75,13.12)
}
\rput{90}(67.5,14.38){\psellipse[fillstyle=solid](0,0)(1.57,-1.56)}
\psline(67.5,10)
(67.5,0)
(67.5,9.38)(67.5,13.12)
\end{pspicture}
\caption{$f$ is a function if and only if it is a comonoid homomorphism}
\label{fig-comonoid}
\end{center}
\end{figure}

\end{document}